\documentclass[letterpaper]{article} 
\usepackage{aaai25}  
\usepackage{times}  
\usepackage{helvet}  
\usepackage{courier}  
\usepackage[hyphens]{url}  
\usepackage{graphicx} 
\usepackage{natbib}  
\usepackage{caption} 
\usepackage{algorithm}
\usepackage{algpseudocode}
\usepackage{amsfonts}       
\usepackage{amsmath}
\usepackage{amssymb}
\usepackage{multirow}
\usepackage[table,xcdraw]{xcolor}
\usepackage{newfloat}
\usepackage{listings}
\DeclareCaptionStyle{ruled}{labelfont=normalfont,labelsep=colon,strut=off} 
\floatstyle{ruled}
\newfloat{listing}{tb}{lst}{}
\floatname{listing}{Listing}
\title{Topology-Aware 3D Gaussian Splatting: \\Leveraging Persistent Homology for Optimized Structural Integrity}
\author{
    Tianqi Shen\textsuperscript{\rm 1,\rm 4},
    Shaohua Liu\textsuperscript{\rm 2,\rm 3},
    Jiaqi Feng\textsuperscript{\rm 2},
    Ziye Ma\textsuperscript{\rm 1},
    Ning An\textsuperscript{\rm 4,\rm 5}\thanks{Corresponding author: Ning An (ning.an@ccteg-bigdata.com)}
}
\affiliations{
    \textsuperscript{\rm 1}Department of Computer Science, City University of Hong Kong\\
    \textsuperscript{\rm 2}Image Processing Center, Beihang University\\
    \textsuperscript{\rm 3}Shen Yuan Honors College, Beihang University\\
    \textsuperscript{\rm 4}Research Institute of Mine Artificial Intelligence, China Coal Research Institute\\
    \textsuperscript{\rm 5}State Key Laboratory of Intelligent Coal Mining and Strata Control\\
    tianqshen5-c@my.cityu.edu.hk, liushaohua@buaa.edu.cn, fengjiaqi@buaa.edu.cn,\\
    ziyema@cityu.edu.hk, ning.an@ccteg-bigdata.com
}

\usepackage{bibentry}

\begin{document}

\maketitle

\begin{abstract}
Gaussian Splatting (GS) has emerged as a crucial technique for representing discrete volumetric radiance fields. It leverages unique parametrization to mitigate computational demands in scene optimization. This work introduces Topology-Aware 3D Gaussian Splatting (Topology-GS), which addresses two key limitations in current approaches: compromised pixel-level structural integrity due to incomplete initial geometric coverage, and inadequate feature-level integrity from insufficient topological constraints during optimization. To overcome these limitations, Topology-GS incorporates a novel interpolation strategy, Local Persistent Voronoi Interpolation (LPVI), and a topology-focused regularization term based on persistent barcodes, named PersLoss. LPVI utilizes persistent homology to guide adaptive interpolation, enhancing point coverage in low-curvature areas while preserving topological structure. PersLoss aligns the visual perceptual similarity of rendered images with ground truth by constraining distances between their topological features. Comprehensive experiments on three novel-view synthesis benchmarks demonstrate that Topology-GS outperforms existing methods in terms of PSNR, SSIM, and LPIPS metrics, while maintaining efficient memory usage. This study pioneers the integration of topology with 3D-GS, laying the groundwork for future research in this area.
\end{abstract}

%
\begin{links}
	\link{Code}{https://github.com/AmadeusSTQ/Topology-GS}
\end{links}

\section{Introduction}
Novel-view synthesis (NVS) is a critical area in computer vision and graphics, aimed at generating new views of a scene from a limited set of input images \cite{yan2021deep}. Significant advancements in this area have been driven by techniques such as Neural Radiance Fields (NeRF) \cite{mildenhall2021nerf, zhang2020nerf++, barron2021mip, xu2023grid} and 3D Gaussian Splatting (3D-GS) \cite{kerbl3Dgaussians, yu2024mip, li2024dngaussian, guedon2024sugar}. NeRF employs an implicit representation that models a scene as a continuous volumetric field, optimized through neural networks. Despite its impressive results, NeRF is computationally intensive due to volumetric ray-marching \cite{fridovich2022plenoxels, muller2022instant}.
In contrast, 3D-GS offers an efficient point-based representation by utilizing Gaussian primitives (hereafter referred to as Gaussians), parametrized by their positions $\boldsymbol{\mu}$ and covariance matrices $\boldsymbol{\Sigma}$, along with learnable attributes for color $\boldsymbol{c}$ and opacity $\boldsymbol{o}$. Combined with a tile-based rasterization pipeline, 3D-GS achieves high-fidelity rendering results and real-time rendering speeds \cite{wu2024recent}.

The density function of a 3D Gaussian in 3D-GS is defined as:
\begin{equation}
	f(\boldsymbol{x}\mid\boldsymbol{\mu},\boldsymbol{\Sigma})=
	\operatorname{exp}{\{-\frac{1}{2}(\boldsymbol{x}-\boldsymbol{\mu})^T\boldsymbol{\Sigma}^{-1}(\boldsymbol{x}-\boldsymbol{\mu})\}},
\end{equation}
where $\boldsymbol{x}$ is an arbitrary 3D point. 
To ensure the covariance matrix is positive semi-definite, \cite{kerbl3Dgaussians} proposed decomposing it using rotation and scaling: $\boldsymbol{\Sigma} = \boldsymbol{R} \boldsymbol{S} \boldsymbol{S}^T \boldsymbol{R}^T$, with the rotation represented using quaternions. 

Finally, the 3D Gaussians are projected onto a 2D image, and their corresponding 2D pixel colors are determined by blending the $N$ ordered Gaussians at the queried pixel as:
\begin{equation}
	\boldsymbol{C}=\sum\nolimits_{i=1}^N {o_i \prod\nolimits_{j=1}^{i-1} {(1-o_j)\boldsymbol{c}_i}},
\end{equation}
where $\boldsymbol{c}_i$ and $o_i$ represent the color and opacity of Gaussian $i$, respectively.
\begin{figure*}[t]
	\includegraphics[width=17.82cm]{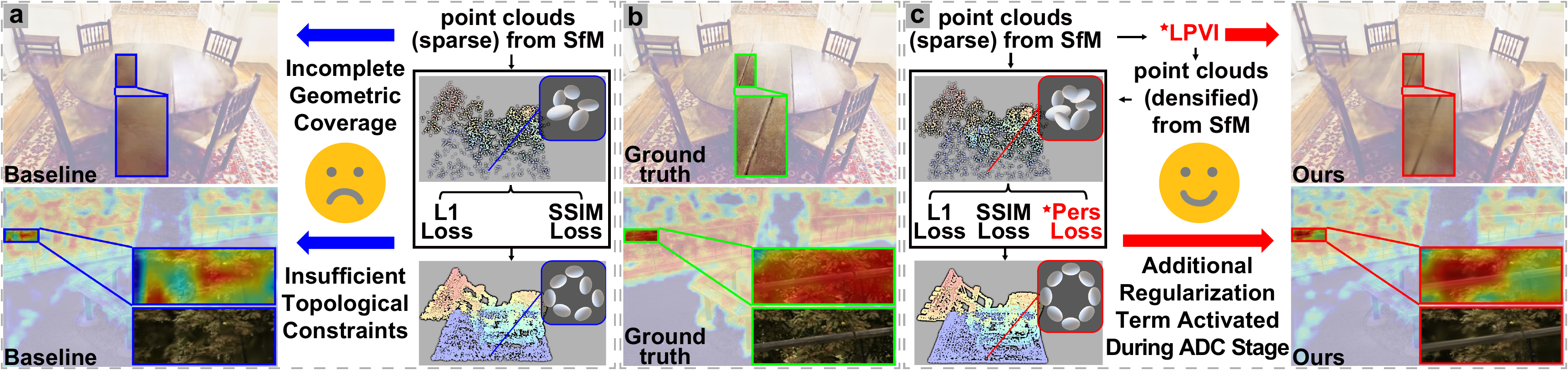}
	\captionof{figure}{
		Illustration of main motivations and our contributions. 
		\textbf{a} Typical pipeline of state-of-the-art (SOTA) methods and visualization of their two main limitations.
		\textbf{b} Visualization of the ground truth.
		\textbf{c} Pipeline of our method (i.e., Topology-GS) and visualization showing improvements over existing methods.
		Feature maps, visualized using EigenCAM \cite{muhammad2020eigen}, are from the last layer of VGG16 \cite{simonyan2014very}, one of the layers used to compute the LPIPS \cite{zhang2018unreasonable} metric. The differences in point clouds \& Gaussians before and after training for both the baseline and our method are illustrated using Lego from \cite{mildenhall2021nerf} as a conceptual example.
	}
	\label{fig_intro}
\end{figure*}

Typically, 3D-GS initializes \textit{plain} Gaussian structures with a \textit{sparse} point cloud generated by Structure from Motion (SfM) \cite{schoenberger2016sfm}, a photogrammetric technique that reconstructs 3D structures from sequences of 2D images by matching features across multiple images. During optimization, an Adaptive Density Control (ADC) mechanism \cite{kerbl3Dgaussians} manages the growth or pruning of each Gaussian \textit{independently} to model \textit{complex} structures. As illustrated in Figure \ref{fig_intro}, these design choices present two primary limitations that hinder the rendered images from accurately reflecting the structural integrity of the scenes. Here, structural integrity refers to \textit{the consistency of geometric shapes and topological features throughout the 3D reconstruction or image rendering process}, ensuring the preservation of local texture patterns and global semantic relationships. Below, we provide a detailed analysis of these limitations and propose solutions using Persistent Homology (PH) \cite{huber2021persistent} from topological data analysis (TDA) \cite{chazal2021introduction}.

Firstly, since SfM relies heavily on the density and quality of feature matches, which can fluctuate substantially across different image regions, the resulting point clouds often exhibit notably sparse \textit{geometric coverage} in areas with low curvature \cite{ververas2025sags}. 
As shown in Figure \ref{fig_intro} (upper left, blue arrow), this sparsity compromises 3D-GS rendering quality with the loss of planar texture details, undermining \textbf{pixel-level structural integrity}.
Furthermore, the ADC mechanism in 3D-GS can cause significant Gaussian displacement in highly sparse areas \cite{ververas2025sags}. Similarly, other point cloud enhancement methods \cite{niemeyer2024radsplat, foroutan2024does, lee2025deblurring, ververas2025sags} also carry a common risk of disrupting the topological structure.

To tackle the specific challenge in 3D-GS, we propose a novel approach called Local Persistent Voronoi Interpolation (LPVI). 
LPVI introduces a unique TopoDiff check, which leverages differences in persistent homology as a guiding criterion. This mechanism enables adaptive switching between preserving topological structures in 3D space and performing interpolation in 2D space, effectively mitigating the sparsity issue in low-curvature regions.
By adaptively increasing point coverage in these areas, LPVI enhances surface detail without compromising overall structural integrity. As shown in Figure \ref{fig_intro} (upper right, red arrow), the interpolation process significantly improves the rendering of low-curvature surfaces, recovering previously missing textures and ultimately refining pixel-level structural consistency.

Secondly, the lack of \textit{topological constraints} during training results in poor \textbf{feature-level structural integrity}. 
As shown in Figure \ref{fig_intro} (lower left, indicated by the blue arrow), existing methods often use pixel-level loss functions (L1 loss and SSIM loss) to guide the independent optimization of Gaussians. 
This results in high PSNR and SSIM \cite{wang2004image} metrics but also high LPIPS \cite{zhang2018unreasonable} metrics, which measure visual perceptual similarity by calculating the distances between the feature maps of the rendered and ground truth images. 
Some alternative approaches \cite{lu2024scaffold, yang2024spectrally, turkulainen2024dn, guedon2024sugar} impose structural constraints on Gaussians but do not ensure the reduction of distances between feature maps of rendered images, while relying on deeper networks to reduce LPIPS is not practical for lightweight applications.

To address the specific challenge in 3D-GS, we introduce a topology-based regularization term inspired by persistent homology, termed PersLoss. 
This term assesses the persistent homology of both rendered and ground truth images by analyzing their truncated persistence barcodes, which represent the birth and death times of topological features with top-k lifespans, with its differentiability enabled by techniques in \cite{gabrielsson2020topology, zhang2022convergence}.
Activated during the ADC phase, it ensures the alignment of 2D rendered images with the abstract features of the ground truth, directly guiding the visual perceptual similarity between rendered images and the ground truth. 
As shown in Figure \ref{fig_intro} (lower right, red arrow), post-training, the rendered images exhibit improved structural \& semantic information and reduced LPIPS, enhancing overall visual quality without incurring extra computational overhead during rendering.

As illustrated in Figure \ref{fig_intro}\&\ref{fig_result}, extensive experiments across three NVS benchmarks demonstrate that our method (i.e. Topology-GS), which incorporates the proposed LPVI and PersLoss, significantly enhances PSNR and SSIM metrics, while reducing LPIPS. This study represents the first successful integration of topology with 3D-GS, highlighting a substantial synergy between the two.

\begin{figure*}[t]
	\centering
	\includegraphics[width=17.8cm]{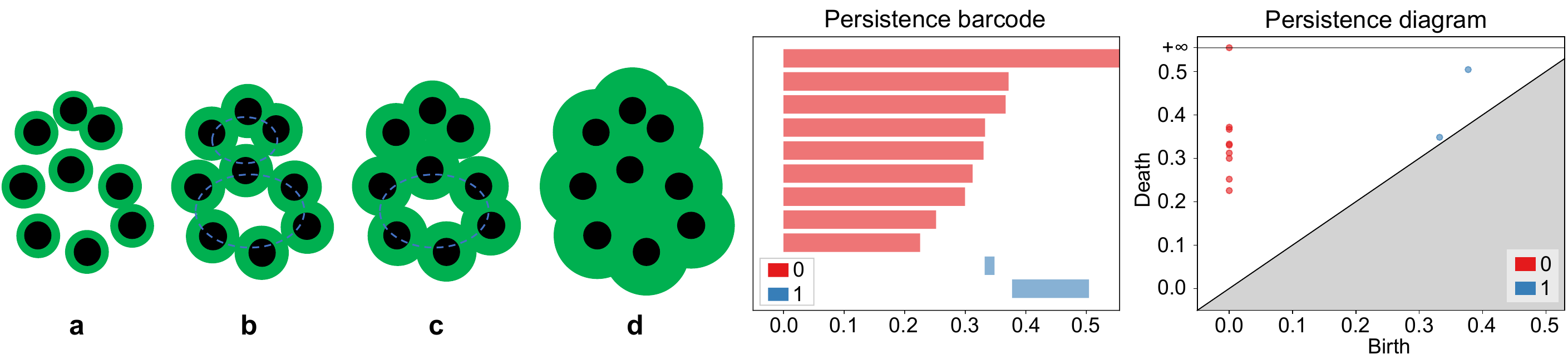}
	\caption{An illustration of persistent homology. Panels a to d depict the evolution of the topological structure when a filtration based on the Euclidean distance (indicated in green) is applied to a set of black points. The two images on the right display the corresponding persistence barcode and persistence diagram. Blue bars or points represent 0-dimensional homology, i.e., connected components, while red bars or points represent 1-dimensional homology, i.e., loops.
	}
	\label{fig_ph}
\end{figure*}

\subsection{Notations}
The rendered image obtained from 3D-GS is denoted by $\boldsymbol{y} \subset \mathbb{R}^{3\times H \times W}$, where $H$ represents the height and $W$ represents the width of the image. Correspondingly, the ground truth image is represented by $\hat{\boldsymbol{y}} \subset \mathbb{R}^{3\times H \times W}$.
Let $Vor^{nD}(x)$ denote the vertices of the $n$-dimensional Voronoi region of a point $x$. The symbol $H^k$ represents the $k$-dimensional homology group, which quantifies various topological features such as connected components ($H^0$), loops ($H^1$), and voids ($H^2$). The Betti numbers, denoted by $\beta^i$, indicate the rank of the $i$-th homology group, reflecting the count of $k$-dimensional features.
The operation $PD(\cdot)$ denotes the process of obtaining the persistence diagram from a three-dimensional point set. The scalar-valued function $f_W$ is a filter function with model parameters $W$.

\section{Related Works}
\subsection{NeRF and 3D-GS} 
\label{subsection:nvs}
NeRF \cite{mildenhall2021nerf} uses a Multi-Layer Perceptron (MLP) to represent the radiance field and volume rendering for realistic images. However, it faces high computational costs, long training times, and slow rendering speeds due to volumetric ray-marching. Methods like Plenoxels \cite{fridovich2022plenoxels} and InstantNGP \cite{muller2022instant} aim to address these issues by using sparse voxel grids and hash-grid encodings, respectively, but the need for sample queries still hampers rendering speed.

3D-GS extends NeRF by incorporating Gaussians with learnable attributes and a tile-based rasterization pipeline, enabling faster and higher-fidelity rendering. While advancements have been achieved in areas such as anti-aliasing \cite{yu2024mip,song2024sa}, high-frequency signal modeling \cite{yang2024spec}, few-shot view synthesis \cite{zhu2025fsgs,li2024dngaussian}, and 3D mesh reconstruction \cite{guedon2024sugar}, two main challenges persist.

\textit{Firstly}, the quality of the initial point cloud coverage significantly impacts the reconstruction quality. However, the ADC mechanism, which aims to mitigate this issue, often introduces significant Gaussian displacement in sparse areas \cite{ververas2025sags}. Existing solutions, such as optimizing NeRF as a prior \cite{niemeyer2024radsplat, foroutan2024does}, expanding point clouds \cite{lee2025deblurring}, and curvature-based enhancements \cite{ververas2025sags}, risk disrupting the topology of the scene. To address this, we introduce LPVI, a novel approach that refines sparse coverage while preserving the topology of the scene.
\textit{Secondly}, the heuristic optimization of 3D-GS often neglects scene structural constraints, leading to poor feature-level integrity. While integrating with deep neural networks can improve ground truth similarity, it comes at the cost of computational efficiency \cite{zou2024triplane}. Recent approaches, such as hierarchical scene representation \cite{lu2024scaffold}, graph-based structural modeling \cite{yang2024spectrally}, and geometric priors \cite{turkulainen2024dn}, have improved the structural quality of the reconstructed scenes. However, these methods do not directly guide semantic feature similarity. To bridge this gap, we develop PersLoss, a novel topological constraint that optimizes feature map distances, enhancing visual perceptual similarity, which represents a semantic-level similarity.

\begin{figure*}[t]
	\centering
	\includegraphics[width=17.8cm]{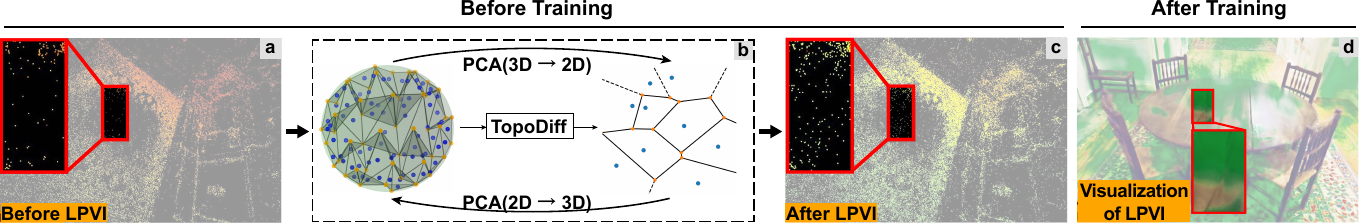}
	\caption{An illustration of LPVI. Figure b shows the principle of LPVI, which involves switching between 3D and 2D local Voronoi interpolation based on topological differences. Figures a and c show point clouds of the table, with a displaying the initial sparse point cloud and c showing the point cloud enhanced by LPVI. 
		Figure d presents the final rendered result, where the green shading visualizes the effect of the interpolated point cloud rendered onto the planar image.
	}
	\label{fig_LPVI}
\end{figure*}

\subsection{Topological Data Analysis}
Recent advances in topological data analysis (TDA) have enhanced topology-based machine learning techniques \cite{chazal2021introduction}. Persistent homology (PH) \cite{huber2021persistent}, a key method in TDA, captures topological features like "holes" in data, which are invariant to noise and scale variations, and analogous to high-dimensional abstract features in topological neural networks \cite{reinauer2021persformer, de2022ripsnet, nishikawa2024adaptive}. In our study, using alpha complexes is suitable for point cloud representation \cite{difrancesco2020implications}.

An alpha complex is derived from the Delaunay triangulation \cite{preparata2012computational} of a point set, filtered by a parameter $\alpha$. For points $P \subset \mathbb{R}^d$, the Delaunay triangulation yields simplices included in the alpha complex if circumscribed by a sphere of radius $\sqrt{\alpha}$. The nested sequence of simplicial complexes forms a filtration:
\begin{equation}
	\emptyset=K_0\subseteq K_1 \subseteq \dots \subseteq K_n=K,
\end{equation}
where each $K_i$ relates to an alpha complex for some $\alpha_i$.

As illustrated in Figure \ref{fig_ph}, PH tracks the birth and death of topological features as $\alpha$ changes, visualized through persistence diagrams (PD) and persistence barcodes (PB). The PD, a scatter plot, maps each point $(\alpha_{birth}, \alpha_{death})$ to the birth and death of a topological feature. Points further from the diagonal $\alpha_{birth} = \alpha_{death}$ indicate more significant features. The PB consists of a series of intervals, each representing the lifespan of a topological feature. Longer bars denote features that endure across a broad range of $\alpha$ values, denoting their significance, while shorter bars typically suggest noise.

PH has facilitated various tasks \cite{pun2022persistent, zia2024topological}, integrating topological features into machine learning models to capture data structures \cite{wang2020topogan, hu2024topology}, and employing PH-based loss functions to train neural networks preserving topological characteristics \cite{hu2019topology, clough2020topological, qi2023dynamic}.
Incorporating TDA into 3D-GS, as shown in this work, leverages PH to preserve the structural integrity of point clouds during the initialization and optimization of Gaussians. 
Recent studies \cite{jignasu2024sdfconnect, nishikawa2024adaptive} highlight the potential of PH in this context, demonstrating its effectiveness in capturing the structural properties of point clouds. Building on these advancements, our approach further enhances the rendering quality of 3D scene representations.

\section{Methodology}
\subsection{Local Persistent Voronoi Interpolation}
In this study, we propose the Local Persistent Voronoi Interpolation (LPVI), as illustrated in Figure \ref{fig_LPVI}.b, which visually demonstrates the mechanism of LPVI. This algorithm interpolates sparse 3D point clouds while preserving the local topological structure by adaptively switching between 3D and 2D Voronoi Interpolation based on changes in topological features before and after interpolation. The pseudocode is presented in Algorithm \ref{alg:LPVI}. To further illustrate the ideas behind LPVI, we break down the procedures in Algorithm \ref{alg:LPVI} and explain them step by step.

\paragraph{3D Voronoi Interpolation}
The execution of LPVI begins by defining an empty set for processed point indices and another for interpolated points. The main loop iterates over each point $x_l$ in the initial sparse point cloud, processing each point only once. For each unprocessed point, the algorithm identifies its $K$-nearest neighbors and performs a 3D Voronoi tessellation \cite{tanemura1983new}. The vertices of this tessellation, shown as the orange points in Figure \ref{fig_LPVI}.b, are potential candidates for interpolation.

\paragraph{Topological Assessment}
A critical step in the algorithm is the evaluation of topological changes, facilitated by the $\textit{TopoDiff}$ function, which computes the Wasserstein distance $W\_Dist$ between PDs of two point sets $\mathcal{X}_l$ and $\mathcal{\hat{X}}_l$:
\begin{equation}
	\textit{W\_Dist}(\textit{PD}(\mathcal{X}_l), \textit{PD}(\mathcal{\hat{X}}_l)),
\end{equation}
where $\mathcal{X}_l$ is the $K$-nearest neighbor set centered at $x_l$, and $\mathcal{\hat{X}}_l$ is this set plus the interpolation points from the 3D Voronoi tessellation.

\begin{algorithm}
	\caption{Local Persistent Voronoi Interpolation (LPVI)}
	\label{alg:LPVI}
	\begin{algorithmic}[1]
		\Require $\mathcal{X} = \{ x_1, \cdots, x_m \} \subset \mathbb{R}^3$: a sparse point set
		\Require $K$: the number of neighbors for 3D Voronoi
		\Require $K'$: the number of neighbors for 2D Voronoi
		\Require $\tau$: the threshold of topological differences
		\Ensure $\hat{\mathcal{X}} \subset \mathbb{R}^3$: an interpolated point set 
		\State $I \leftarrow \emptyset$, $\hat{\mathcal{X}} \leftarrow \emptyset$, $\mathcal{M} = \{1, \cdots, m\} \leftarrow$ indices of $\mathcal{X}$
		\For{$l \leftarrow 1$ \textbf{to} $m$}
		\If{$l \notin I$}
		\State $I \leftarrow I \cup \{ l \}$
		\State $\{ \mu_1, \cdots, \mu_K \}_{\mu_i \in \mathcal{M}} \leftarrow$ the indices of the $K$ nearest neighbors of $x_l \in \mathcal{X}$
		\State $\{ v_1, \cdots, v_p \} \leftarrow$ $Vor^{3D}(x_l)$
		\State $\mathcal{X}_l \leftarrow \{ x_l, x_{\mu_1}, \cdots, x_{\mu_K} \}$
		\State $\mathcal{\hat{X}}_l \leftarrow \{ x_l, x_{\mu_1}, \cdots, x_{\mu_K}, v_1, \cdots, v_p \}$
		\If{$\textbf{TopoDiff}(\mathcal{X}_l, \mathcal{\hat{X}}_l) < \tau$}
		\State $I \leftarrow I \cup \{ \mu_1, \cdots, \mu_K \}$
		\State $\hat{\mathcal{X}} \leftarrow \hat{\mathcal{X}} \cup \{v_1, \cdots, v_p \}$ 
		\State $\textbf{continue}$
		\EndIf
		\State $\{ \lambda_1, \cdots, \lambda_{K'} \}_{\lambda_i \in \mathcal{M}} \leftarrow$ the indices of the $K'$ nearest neighbors of $x_l \in \mathcal{X}$
		\State $\{ x_0', x_1', \cdots, {x'}_{K'} \}, U\leftarrow$ Dimensionality Reduction Using $PCA_{3\rightarrow 2}(\{ x_l, x_{\lambda_1}, \cdots, x_{\lambda_{K'}} \})$
		\State $\{ v_1', \cdots, v_{p'}' \} \leftarrow$ $Vor^{2D}(x_0')$
		\State $\{ \hat{x}_1, \cdots, \hat{x}_{p'} \} \leftarrow \{ \hat{x}_i = U v_i' + x_l \}_{i = 1, \cdots, p'}$ 
		\State $I \leftarrow I \cup \{ \lambda_1, \cdots, \lambda_{K'} \}$
		\State $\hat{\mathcal{X}} \leftarrow \hat{\mathcal{X}} \cup \{ \hat{x}_1, \cdots, \hat{x}_{p'} \}$
		\EndIf
		\EndFor
	\end{algorithmic}
\end{algorithm}

The Wasserstein distance is a measure of the difference between two probability distributions, capturing both the magnitude and the geometry of the changes in topological features. In this paper, we adopt the Wasserstein distance as defined in \cite{maria2014gudhi,kerber2017geometry}.
If the topological difference remains below a predefined threshold $\tau$, the vertices are added to the interpolated set. Conversely, a significant topological change triggers a switch to 2D interpolation.

\paragraph{2D Voronoi Interpolation}
In cases requiring 2D interpolation, the algorithm selects a smaller set of neighbors, applies Principal Component Analysis (PCA) to project these points onto a plane, and performs a 2D Voronoi tessellation. The vertices from this tessellation are then mapped back to the 3D space using the principal components (step 17 in Algorithm \ref{alg:LPVI}), ensuring the preservation of local geometric structures. Here, the notation $PCA_{3\rightarrow 2}$ refers to the process of reducing dimensionality from 3D to 2D using PCA \cite{yamada2020inferring}, yielding the reduced point set and the first two principal vectors, denoted by $U = [u_1, u_2]$.

We utilize 3D-to-2D Voronoi Interpolation for several reasons. First, manifold learning fundamentally assumes high-dimensional data lie on a low-dimensional manifold \cite{fefferman2016testing}, with the minimal neighborhood of any non-isolated point approximated by a tangent hyperplane (or tangent plane for 3D point clouds). Second, point clouds from SfM are often sparse on planar surfaces \cite{ververas2025sags}, making 2D embedding better suited for interpolating low-curvature regions. Third, embedding in low-dimensional manifolds is computationally more efficient for interpolation than in high dimensions \cite{melodia2020persistent}.

\subsection{Persistent Homology Loss Function}
As outlined in the introduction, NVS evaluation metrics include PSNR, SSIM, and LPIPS. Unlike PSNR and SSIM, LPIPS evaluates perceptual similarity by comparing semantic feature maps extracted from a pretrained vision network. It measures the feature-level structural integrity of the scene, where smaller distances indicate higher rendering quality. 

To maintain efficiency, we avoid using deeper networks in 3D-GS, as they would increase memory usage and reduce rendering speed. Instead, we propose a persistent homology-based loss function, $\operatorname{PersLoss}$, as an additional term in the total loss to guide the Gaussians in learning abstract structural and semantic features. The computation of our loss function involves the following steps:

\paragraph{2D Image and 3D RGB Space} After the rasterization step of 3D-GS, we obtain the rendered image $\boldsymbol{y}$, along with the ground truth image $\hat{\boldsymbol{y}}$. Given that the true ground truth for 3D point clouds (or 3D Gaussians) is unknown, we do not utilize them for subsequent persistent homology calculation. Instead, we rely on the rendered image obtained from differentiable rasterization, which has a known ground truth, for loss calculation. The gradient information from the 2D image propagates back through the differentiable rasterization of 3D-GS, optimizing the training parameters.

To facilitate persistent homology calculations, we reshape both images:
\begin{equation}
	\boldsymbol{y'}=S(\boldsymbol{y}), \quad \hat{\boldsymbol{y}}'=S(\hat{\boldsymbol{y}}),
\end{equation}
where $S$ reshapes both $\boldsymbol{y}$ and $\hat{\boldsymbol{y}}$ to the size of $HW \times 3$.

This transformation treats the 2D image data as data in a 3D RGB space, allowing us to perform persistent homology calculations in this space.

\paragraph{Filtering and Truncated Persistent Barcode}
We utilize the alpha complex to filter $\boldsymbol{y'}$ and $\hat{\boldsymbol{y}}'$ for PH computation, obtaining topological features expressed in the form of PD and PB. For each feature, its birth and death values are denoted as $b^k$ and $d^k$, respectively. The lifespan of a feature in the rendered image is represented as $(b^k, d^k)$, while the lifespan of a feature in the ground truth image is represented as $(\hat{b}^k, \hat{d}^k)$. In the context of PB, longer bars indicate features that persist over a wide range of alpha values, signifying their importance, while shorter bars often correspond to noise. Consequently, we set three parameters, $k_0$, $k_1$, and $k_2$, to select the top-$k$ longest lifespans of topological features for 0-dimensional, 1-dimensional, and 2-dimensional homology, respectively. We then obtain the truncated PBs for both images, denoted as $pb$ and $\hat{pb}$:
\begin{equation}
	pb = 
	\left\{
	\begin{array}{l}
		(b^0_1, d^0_1), \cdots, (b^0_{k_0}, d^0_{k_0}) \\
		(b^1_1, d^1_1), \cdots, (b^1_{k_1}, d^1_{k_1}) \\
		(b^2_1, d^2_1), \cdots, (b^2_{k_2}, d^2_{k_2})
	\end{array},
	\right.
\end{equation}
\begin{equation}
	\hat{pb} = 
	\left\{
	\begin{array}{l}
		(\hat{b}^0_1, \hat{d}^0_1), \cdots, (\hat{b}^0_{k_0}, \hat{d}^0_{k_0}) \\
		(\hat{b}^1_1, \hat{d}^1_1), \cdots, (\hat{b}^1_{k_1}, \hat{d}^1_{k_1}) \\
		(\hat{b}^2_1, \hat{d}^2_1), \cdots, (\hat{b}^2_{k_2}, \hat{d}^2_{k_2})
	\end{array}.
	\right.
\end{equation}

Since we are only interested in high-dimensional features, we discard the topological features beyond the top-$k$.

\paragraph{PersLoss}
We then calculate the total difference between the birth and death values of the features in each topological dimension to serve as our regularization term:
\begin{align}
	&\operatorname{PersLoss} = \nonumber \\
	&\sum_{i=0}^2 {\frac{\beta^i}{\sum_{k=0}^2 {\beta^k}} \sum_{j=1}^{k_i} \left(\left|b^i_j - \hat{b}^i_j\right|^2+\left|d^i_j - \hat{d}^i_j\right|^2\right)},
\end{align}
where $\beta^i$ represents the Betti numbers, indicating the number of $i$-dimensional features.

Finally, our topology-aware total loss function comprises a standard supervision loss term, $L_{\text{supv}}$, which combines L1 loss and SSIM loss, and a topology-focused regularization term, $\operatorname{PersLoss}$. The regularization term quantifies the similarity between the predicted topology and the desired topology. As discussed in the introduction, both terms are differentiable, allowing gradients from the topology-aware total loss to propagate through the entire pipeline. This effectively constrains and guides the optimization of Gaussian parameters, thereby improving the feature-level structural integrity of the rendered output.

\begin{table*}[t]
	\caption{RENDERING QUALITY ON MIP-NERF360, TANKS\&TEMPLES AND DEEP BLENDING}
	\label{tab_requ}
	\centering
	\resizebox{1.0\linewidth}{!} 
	{
		\begin{tabular}{cccccccccccc}
			\hline
			\multirow{3}{*}{\begin{tabular}[c]{@{}c@{}}Dataset\\ Metrics\\ Method\end{tabular}} & \multirow{3}{*}{Year} & \multirow{3}{*}{Paper} & \multicolumn{3}{c}{Mip-NeRF360}                                                                                                                                                                            & \multicolumn{3}{c}{Tanks \& Temples}                                                                                                                                                                       & \multicolumn{3}{c}{Deep Blending}                                                                                                                                                                          \\ \cline{4-12} 
			&                       &                        & \multirow{2}{*}{\begin{tabular}[c]{@{}c@{}}PSNR\\ ↑\end{tabular}} & \multirow{2}{*}{\begin{tabular}[c]{@{}c@{}}SSIM\\ ↑\end{tabular}} & \multirow{2}{*}{\begin{tabular}[c]{@{}c@{}}LPIPS\\ ↓\end{tabular}} & \multirow{2}{*}{\begin{tabular}[c]{@{}c@{}}PSNR\\ ↑\end{tabular}} & \multirow{2}{*}{\begin{tabular}[c]{@{}c@{}}SSIM\\ ↑\end{tabular}} & \multirow{2}{*}{\begin{tabular}[c]{@{}c@{}}LPIPS\\ ↓\end{tabular}} & \multirow{2}{*}{\begin{tabular}[c]{@{}c@{}}PSNR\\ ↑\end{tabular}} & \multirow{2}{*}{\begin{tabular}[c]{@{}c@{}}SSIM\\ ↑\end{tabular}} & \multirow{2}{*}{\begin{tabular}[c]{@{}c@{}}LPIPS\\ ↓\end{tabular}} \\
			&                       &                        &                                                                   &                                                                   &                                                                    &                                                                   &                                                                   &                                                                    &                                                                   &                                                                   &                                                                    \\ \hline
			3D-GS \cite{kerbl3Dgaussians}                                                                               & 2023                  & TOG                    & 27.21                                                             & 0.815                                                             & 0.214                                                              & 23.14                                                             & 0.841                                                             & 0.183                                                              & 29.41                                                             & 0.903                                                             & \textbf{0.243}                                                     \\
			Scaffold-GS \cite{lu2024scaffold} (\textbf{baseline})                                                              & 2024                  & CVPR                   & 28.84                                                             & 0.848                                                             & 0.220                                                              & 23.96                                                             & 0.853                                                             & 0.177                                                              & \underline{30.21}                                                 & \underline{0.906}                                                 & 0.254                                                              \\
			Mip-Splatting \cite{yu2024mip}                                                                       & 2024                  & CVPR                   & 27.79                                                             & 0.827                                                             & 0.203                                                              & 23.65                                                             & 0.849                                                             & 0.211                                                              & 29.68                                                             & 0.903                                                             & 0.309                                                              \\
			2D-GS \cite{huang20242d}                                                                               & 2024                  & SIGGRAPH               & 28.75                                                             & 0.870                                                             & 0.213                                                              & 22.96                                                             & 0.825                                                             & 0.217                                                              & 29.49                                                             & 0.899                                                             & 0.259                                                              \\
			Revised-GS \cite{rota2025revising}                                                                          & 2024                  & ECCV                   & 27.70                                                             & 0.823                                                             & 0.223                                                              & \underline{24.10}                                                 & \underline{0.857}                                                 & 0.183                                                              & 29.64                                                             & 0.905                                                             & 0.303                                                              \\
			Pixel-GS \cite{zhang2025pixelgs}                                                                            & 2024                  & ECCV                   & \underline{29.11}                                                 & \underline{0.872}                                                 & \textbf{0.165}                                                     & 23.61                                                             & 0.851                                                             & \underline{0.161}                                                  & 28.83                                                             & 0.892                                                             & 0.251                                                              \\
			Topology-GS (\textbf{ours})                                                         & 2025                  & AAAI                   & \textbf{29.50}                                                    & \textbf{0.874}                                                    & \underline{0.179}                                                  & \textbf{24.26}                                                    & \textbf{0.860}                                                    & \textbf{0.160}                                                     & \textbf{30.45}                                                    & \textbf{0.911}                                                    & \underline{0.245}                                                  \\ \hline
		\end{tabular}
	}
	\begin{flushleft}
		\footnotesize
		* \textbf{Bold} indicates the best results, while \underline{underline} represents the second-best results.
	\end{flushleft}
\end{table*}

However, as we theoretically demonstrate in \textit{Section B of the appendix}, the optimization of the topology-aware total loss is prone to fluctuations. Specifically, minimizing $\operatorname{PersLoss}$ may decrease its value, but the optimization of $L_{\text{supv}}$ can perturb the model parameters $W$, causing an increase in the loss value. In our Topology-GS approach, we selectively activate $\operatorname{PersLoss}$ during specific optimization rounds in the ADC stage, which reduces the computational demand of PH and directly guides the adjustment of Gaussians. Nevertheless, due to these fluctuations, we cannot guarantee the convergence of $\operatorname{PersLoss}$ at the end of the ADC stage, making it challenging to ensure that the topological features of the final rendered image approximate those of the ground truth image.

To address this issue, we introduce the following assumptions \cite{zhang2022convergence} and theorem:
\begin{itemize}
	\item \textbf{Assumpt. 0 (A0):} $B=k_0+k_1+k_2$ is sufficiently small such that \textit{Assumptions 1 and 2 in the appendix} hold.
	\item \textbf{Assumpt. 1 (A1):} $f$ is 1-bounded, 1-Lipschitz continuous and 1-Lipschitz smooth with respect to $W$.
	\item \textbf{Assumpt. 2 (A2):} $L_{\text{supv}}(W)$ is $\ell^0$-bounded, $\ell^1$-Lipschitz continuous and $\ell^2$-Lipschitz smooth with respect to $W$.
\end{itemize}
\paragraph{Theorem 1.} \textit{Under Assumptions A0, A1, and A2, and a given stopping condition $\epsilon$, the optimization algorithm using our topology aware total loss stops in $O\left(\frac{1}{\epsilon}\right)$ iterations if step-size $\eta$ is chosen to be:}
\begin{equation}
	\eta \leq \min
	\left\{
	\frac{1}{2\ell^2 + 10\lambda_{\text{topo}}B}, 
	\frac{\epsilon}{4096\lambda_{\text{topo}}^2 B^2}
	\right\}.
\end{equation}

Here, $\lambda_{\text{topo}}$ is the coefficient of $\operatorname{PersLoss}$ in the total loss, and $B$ represents the cardinality of the ground truth PD (excluding the diagonal). \textit{For a detailed proof, see Section C of the appendix.}

This theorem establishes that our topology-aware total loss not only converges despite fluctuations but also converges rapidly with a finite number of training iterations by adjusting hyperparameters. This property makes $\operatorname{PersLoss}$ suitable for use with L1 loss and SSIM loss, as well as the ADC trick, rendering it suitable for 3D-GS tasks.

\section{Experiments}
\textbf{Datasets.} To evaluate the proposed method, consistent with 3D-GS \cite{kerbl3Dgaussians}, we utilized 11 scenes: seven from the Mip-NeRF360 dataset \cite{barron2022mip}, two from the Tanks \& Temples dataset \cite{knapitsch2017tanks}, and two from the Deep Blending dataset \cite{hedman2018deep}.\\
\textbf{Comparison Methods.} We compared the proposed method with state-of-the-art (SOTA) 3D-GS-based approaches for novel-view synthesis, including 3D-GS, Mip-Splatting \cite{yu2024mip}, Scaffold-GS \cite{lu2024scaffold}, and 2D-GS \cite{huang20242d}, as well as the recent Revised-GS \cite{rota2025revising} and Pixel-GS \cite{zhang2025pixelgs}.\\
\textbf{Metrics.} The Topology-GS model was evaluated based on rendering quality, structure preservation (pixels and feature maps), and visual comparisons. Rendering quality was assessed using PSNR, SSIM, and LPIPS metrics.\\
\textit{For additional experimental results, see Section D of the appendix.}

\subsection{Overall Rendering Quality}
Table \ref{tab_requ} presents the rendering quality of various algorithms across three NVS benchmarks. Scaffold-GS serves as the baseline for our comparisons.
Our proposed method, Topology-GS, consistently outperforms Scaffold-GS and nearly all other methods across the three datasets, achieving SOTA results.
The introduction of our LPVI significantly enhances PSNR and SSIM metrics by improving initial geometric coverage. Additionally, incorporating our $\operatorname{PersLoss}$ as a topological constraint leads to a substantial reduction in LPIPS across all datasets. 

\subsection{Structural Preservation}
\begin{figure}[h]
	\centering
	\includegraphics[width=8.3cm]{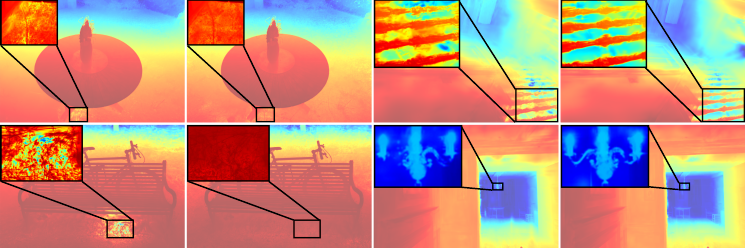}
	\caption{Depth map comparison between a SOTA method (i.e. Scaffold-GS \cite{lu2024scaffold}) and our Topology-GS. Columns 1 and 3 show results from Scaffold-GS, while Columns 2 and 4 show results from Topology-GS. Depth maps are visualized using COLORMAP\_JET.}
	\label{fig_depth}
\end{figure}

As illustrated in Figure \ref{fig_depth}, we rendered depth maps \cite{yuan2025dvp} from perspectives in the test set to visually assess the impact of two proposed enhancements: the LPVI interpolation method and the PersLoss regularization term.

\begin{figure*}[t]
	\centering
	\includegraphics[width=17.8cm]{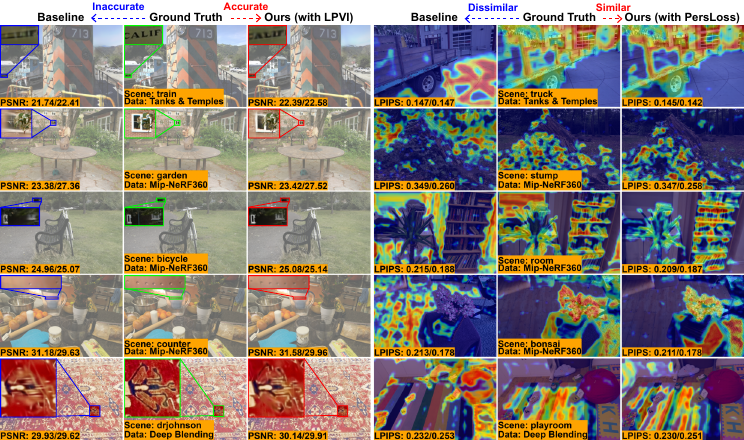}
	\caption{Comparison of Ground Truth (GT) images with those rendered by a SOTA method (i.e. Scaffold-GS \cite{lu2024scaffold}) and our Topology-GS. Columns 2 and 5 show GT images and feature maps; Columns 1 and 4 show Scaffold-GS results; Columns 3 and 6 show Topology-GS results. PSNR and LPIPS scores are annotated, with "/" separating image and scene scores. The rendered images in Column 3 capture low-curvature regions more accurately compared to the Scaffold-GS in Column 1, and the feature maps in Column 6 closely match the GT feature maps in Column 5.}
	\label{fig_result}
\end{figure*}

Comparing Columns 1 and 2, we observe that the baseline (Scaffold-GS) exhibits discontinuous color changes in low-curvature areas on the depth map (e.g., flat ground showing noticeable depth variations). This often indicates significant fluctuations in depth within those regions. In contrast, our method (Topology-GS), enhanced with LPVI, reflects more consistent depth changes in these areas due to the improved pixel-level structural integrity of the initial point cloud in low-curvature regions. Consequently, the depth map's color variations are smoother.
In Columns 3 and 4, the comparison reveals that the baseline model struggles with understanding feature-level structural integrity, particularly in areas with significant structural changes. The baseline's depth map displays sluggish color transitions, failing to adapt to sharp depth variations caused by structural elements (e.g., vague depth regions between railings and gaps). This results in some structures not being properly rendered. However, guided by PersLoss, our method accurately renders these structural changes, ensuring rapid visual transitions in the depth map.

\subsection{Visual Comparison}
Figure \ref{fig_result} presents a visual comparison of Ground Truth (GT) images with rendered images from Scaffold-GS and our proposed method, Topology-GS. 
In the first three columns, GT images are compared with those rendered by Scaffold-GS and Topology-GS. The interpolation introduced in Topology-GS significantly enhances pixel-level details, especially in low-curvature regions, as evidenced by improved PSNR values.
The last three columns show feature map comparisons. The topology-focused regularization term in Topology-GS directs the feature maps to better align with the GT feature maps, resulting in lower LPIPS values and greater perceptual similarity to the GT.

\begin{table*}[t]
	\caption{ABLATION STUDY ON MIP-NERF360, TANKS\&TEMPLES AND DEEP BLENDING}
	\label{tab_abla}
	\centering
	\resizebox{1.0\linewidth}{!} 
	{
		\begin{tabular}{ccccccccccccc}
			\hline
			\multirow{3}{*}{\begin{tabular}[c]{@{}c@{}}Dataset\\ Metrics\\ Ablation\end{tabular}} & \multicolumn{4}{c}{Mip-NeRF360}                                                                                                                                                                                                                                                      & \multicolumn{4}{c}{Tanks \& Temples}                                                                                                                                                                                                                                                 & \multicolumn{4}{c}{Deep Blending}                                                                                                                                                                                                                                                    \\ \cline{2-13} 
			& \multirow{2}{*}{\begin{tabular}[c]{@{}c@{}}PSNR\\ ↑\end{tabular}} & \multirow{2}{*}{\begin{tabular}[c]{@{}c@{}}SSIM\\ ↑\end{tabular}} & \multirow{2}{*}{\begin{tabular}[c]{@{}c@{}}LPIPS\\ ↓\end{tabular}} & \multirow{2}{*}{\begin{tabular}[c]{@{}c@{}}Avg Memory\\ ↓\end{tabular}} & \multirow{2}{*}{\begin{tabular}[c]{@{}c@{}}PSNR\\ ↑\end{tabular}} & \multirow{2}{*}{\begin{tabular}[c]{@{}c@{}}SSIM\\ ↑\end{tabular}} & \multirow{2}{*}{\begin{tabular}[c]{@{}c@{}}LPIPS\\ ↓\end{tabular}} & \multirow{2}{*}{\begin{tabular}[c]{@{}c@{}}Avg Memory\\ ↓\end{tabular}} & \multirow{2}{*}{\begin{tabular}[c]{@{}c@{}}PSNR\\ ↑\end{tabular}} & \multirow{2}{*}{\begin{tabular}[c]{@{}c@{}}SSIM\\ ↑\end{tabular}} & \multirow{2}{*}{\begin{tabular}[c]{@{}c@{}}LPIPS\\ ↓\end{tabular}} & \multirow{2}{*}{\begin{tabular}[c]{@{}c@{}}Avg Memory\\ ↓\end{tabular}} \\
			&                                                                   &                                                                   &                                                                    &                                                                         &                                                                   &                                                                   &                                                                    &                                                                         &                                                                   &                                                                   &                                                                    &                                                                         \\ \hline
			Baseline \cite{lu2024scaffold}                                                                              & 28.84                                                             & 0.848                                                             & 0.220                                                              & 159.98MB                                                                & 23.96                                                             & 0.853                                                             & 0.177                                                              & 76.48MB                                                                 & 30.21                                                             & 0.906                                                             & 0.254                                                              & 54.29MB                                                                 \\
			Baseline+Random                                                                       & 29.15                                                             & 0.859                                                             & 0.206                                                              & 168.35MB                                                                & 24.13                                                             & 0.857                                                             & 0.176                                                              & 109.71MB                                                                & 30.10                                                             & 0.907                                                             & 0.258                                                              & 60.43MB                                                                 \\
			Baseline+DEBLUR \cite{lee2025deblurring}                                                                       & 29.16                                                             & 0.861                                                             & 0.203                                                              & 150.74MB                                                                & 24.23                                                             & 0.861                                                             & 0.174                                                              & 100.06MB                                                                & 30.29                                                             & 0.909                                                             & 0.254                                                              & 55.60MB                                                                 \\
			Baseline+SAGS \cite{ververas2025sags}                                                                         & 29.16                                                             & 0.863                                                             & 0.198                                                              & 147.17MB                                                                & 24.39                                                             & 0.860                                                             & 0.170                                                              & 106.51MB                                                                & 30.38                                                             & 0.910                                                             & 0.251                                                              & 59.01MB                                                                 \\ \hline 
			Baseline+LPVI (\textbf{ours})                                                                  & \underline{29.47}                                                & \underline{0.873}                                                & 0.191                                                              & 173.97MB                                                                & \underline{24.22}                                                & \underline{0.858}                                                & 0.175                                                              & 89.99MB                                                                 & \underline{30.42}                                                & \underline{0.911}                                                & 0.252                                                              & 64.71MB                                                                 \\
			Baseline+PersLoss (\textbf{ours})                                                              & 29.36                                                             & 0.870                                                             & \underline{0.185}                                                 & 160.87MB                                                                & 24.08                                                             & 0.853                                                             & \underline{0.168}                                                 & 76.70MB                                                                 & 30.30                                                             & 0.909                                                             & \underline{0.249}                                                 & 54.62MB                                                                 \\
			Topology-GS (\textbf{ours})                                                                    & 29.50                                                             & 0.874                                                             & 0.179                                                              & \underline{175.36MB}                                                                & 24.26                                                             & 0.860                                                             & 0.160                                                              & \underline{90.66MB}                                                                 & 30.45                                                             & 0.911                                                             & 0.245                                                              & \underline{65.22MB}                                                                 \\ \hline
		\end{tabular}
	}
	\begin{flushleft}
		\footnotesize
		* \underline{Underline} highlights the numbers included in the comparison.
	\end{flushleft}
\end{table*}

\subsection{Ablation Study}
Table \ref{tab_abla} summarizes the results of the ablation study on Topology-GS across three NVS benchmarks. To evaluate the contributions of individual components to the overall performance, we independently removed the LPVI interpolation method and the PersLoss regularization term. Furthermore, for comparison, we replaced LPVI with alternative point cloud densification strategies, including random interpolation (denoted as Random), DEBLUR \cite{lee2025deblurring}, which densifies the point cloud by uniformly sampling within its bounding box, and SAGS \cite{ververas2025sags}, which constructs a graph over low-curvature points and augments the data using their midpoints.

The results indicate that removing the LPVI interpolation method results in a decline in PSNR and SSIM, along with a slight increase in LPIPS, highlighting the importance of LPVI in enhancing pixel-level rendering quality. Similarly, replacing LPVI with alternative interpolation densification methods produces inferior outcomes. Specifically, Random and DEBLUR disrupt the topological structure, whereas SAGS proves effective only for interpolating within low-curvature regions, failing to account for region boundaries. Furthermore, excluding PersLoss leads to a significant increase in LPIPS, reflecting a diminished visual perceptual similarity between the rendered images and the ground truth. These findings underscore the critical role of PersLoss in reducing LPIPS and enhancing the perceptual quality of the rendered images.
\begin{figure}[h]
	\centering
	\includegraphics[width=7.8cm]{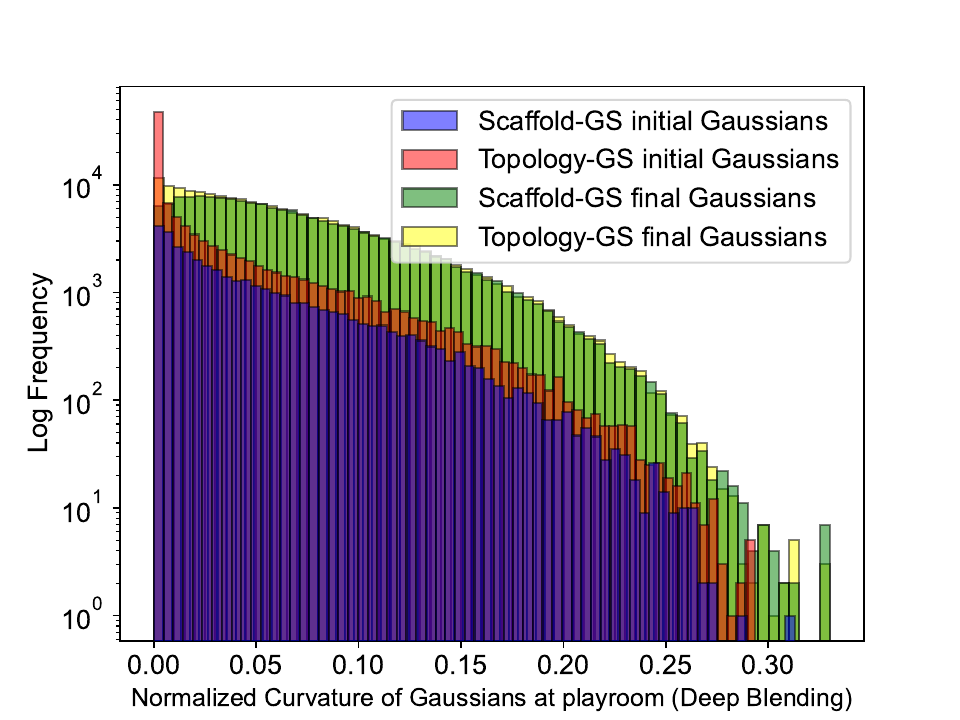}
	\caption{Histogram of Gaussians. Blue and red: the initial point counts for Scaffold-GS (baseline) and Topology-gs, respectively. Green and yellow: the corresponding point counts after training.}
	\label{fig_hist}
\end{figure}

\subsection{Computational Efficiency}
In this section, we discuss the computational overhead of our Topology-GS. As mentioned in the introduction, Topology-GS enhances initial sparse point clouds using our LPVI method and incorporates PersLoss as a structural constraint during training. As shown in Table \ref{tab_abla}, PersLoss acts only during training and does not significantly affect final average memory usage. Therefore, we will primarily discuss the impact of LPVI on final memory usage.

As shown in Table \ref{tab_abla}, while LPVI slightly increases average memory usage, the impact is minimal and remains within an acceptable range for practical applications, with increases of less than 20MB across three benchmarks. Figure \ref{fig_hist} illustrates point counts for Topology-GS (red and yellow) compared to the baseline (blue and green). The red histogram shows increased points in low-curvature areas due to LPVI interpolation. Post-training, the yellow histogram aligns with the baseline in non-low-curvature areas, indicating no disruption elsewhere. Overall, the total points in the yellow histogram remain similar to the green, avoiding additional computational overhead.

\section{Conclusion}
This paper introduced Topology-Aware 3D Gaussian Splatting (Topology-GS), leveraging persistent homology to enhance structural integrity in novel-view synthesis (NVS). By addressing incomplete initial geometric coverage and insufficient topological constraints during training, our approach improves both pixel-level and feature-level integrity. Extensive experiments demonstrate the effectiveness of Topology-GS.
The proposed Topology-GS has the potential to significantly impact the field of NVS by providing a more accurate and efficient method for rendering complex scenes. Its integration of topological methods with 3D Gaussian Splatting could lead to more realistic and detailed renderings, and inspire new approaches to other computer vision tasks.

\section{Acknowledgments}
This work was supported by the National Key Research and Development Program of China (No. 2023YFC2907600), the Key Science and Technology Innovation Project of CCTEG (No. 2024-TD-ZD016-01), and the City University of Hong Kong (\# 9382001). We extend our gratitude to the participants for their time and interest, which has greatly contributed to the research. We are also thankful to Dr. Ali Zia, Postdoctoral Research Fellow at the Australian National University, for his pivotal introduction to topological machine learning, shaping the early stages of our project. Special thanks to Lu Huang, a PhD candidate at Peking University, whose expertise in creating the visual content for this paper was invaluable.

\section{Ethics Statement}
This research did not involve human participants, animals, or the use of sensitive data; therefore, ethical approval was not required.

\bibliography{topology}

\begin{thebibliography}{71}
\providecommand{\natexlab}[1]{#1}

\bibitem[{Barron et~al.(2021)Barron, Mildenhall, Tancik, Hedman,
  Martin-Brualla, and Srinivasan}]{barron2021mip}
Barron, J.~T.; Mildenhall, B.; Tancik, M.; Hedman, P.; Martin-Brualla, R.; and
  Srinivasan, P.~P. 2021.
\newblock Mip-nerf: A multiscale representation for anti-aliasing neural
  radiance fields.
\newblock In \emph{Proceedings of the IEEE/CVF international conference on
  computer vision}, 5855--5864.

\bibitem[{Barron et~al.(2022)Barron, Mildenhall, Verbin, Srinivasan, and
  Hedman}]{barron2022mip}
Barron, J.~T.; Mildenhall, B.; Verbin, D.; Srinivasan, P.~P.; and Hedman, P.
  2022.
\newblock Mip-nerf 360: Unbounded anti-aliased neural radiance fields.
\newblock In \emph{Proceedings of the IEEE/CVF Conference on Computer Vision
  and Pattern Recognition}, 5470--5479.

\bibitem[{Chazal and Michel(2021)}]{chazal2021introduction}
Chazal, F.; and Michel, B. 2021.
\newblock An introduction to topological data analysis: fundamental and
  practical aspects for data scientists.
\newblock \emph{Frontiers in artificial intelligence}, 4: 108.

\bibitem[{Clough et~al.(2020)Clough, Byrne, Oksuz, Zimmer, Schnabel, and
  King}]{clough2020topological}
Clough, J.~R.; Byrne, N.; Oksuz, I.; Zimmer, V.~A.; Schnabel, J.~A.; and King,
  A.~P. 2020.
\newblock A topological loss function for deep-learning based image
  segmentation using persistent homology.
\newblock \emph{IEEE transactions on pattern analysis and machine
  intelligence}, 44(12): 8766--8778.

\bibitem[{Cohen-Steiner, Edelsbrunner, and Harer(2007)}]{cohen2007stability}
Cohen-Steiner, D.; Edelsbrunner, H.; and Harer, J. 2007.
\newblock Stability of persistence diagrams.
\newblock \emph{Discrete \& Computational Geometry}, 37(1).

\bibitem[{Cohen-Steiner et~al.(2010)Cohen-Steiner, Edelsbrunner, Harer, and
  Mileyko}]{cohen2010lipschitz}
Cohen-Steiner, D.; Edelsbrunner, H.; Harer, J.; and Mileyko, Y. 2010.
\newblock Lipschitz functions have L p-stable persistence.
\newblock \emph{Foundations of computational mathematics}, 10(2): 127--139.

\bibitem[{de~Surrel et~al.(2022)de~Surrel, Hensel, Carri{\`e}re, Lacombe, Ike,
  Kurihara, Glisse, and Chazal}]{de2022ripsnet}
de~Surrel, T.; Hensel, F.; Carri{\`e}re, M.; Lacombe, T.; Ike, Y.; Kurihara,
  H.; Glisse, M.; and Chazal, F. 2022.
\newblock RipsNet: a general architecture for fast and robust estimation of the
  persistent homology of point clouds.
\newblock In \emph{Topological, Algebraic and Geometric Learning Workshops
  2022}, 96--106. PMLR.

\bibitem[{Dey and Wang(2022)}]{book111}
Dey, T.; and Wang, Y. 2022.
\newblock \emph{Computational Topology for Data Analysis}.
\newblock Cambridge University Press.
\newblock ISBN 9781009098168.

\bibitem[{DiFrancesco, Bonneau, and
  Hutchinson(2020)}]{difrancesco2020implications}
DiFrancesco, P.-M.; Bonneau, D.; and Hutchinson, D.~J. 2020.
\newblock The implications of M3C2 projection diameter on 3D semi-automated
  rockfall extraction from sequential terrestrial laser scanning point clouds.
\newblock \emph{Remote Sensing}, 12(11): 1885.

\bibitem[{Edelsbrunner and Harer(2010)}]{book222}
Edelsbrunner, H.; and Harer, J. 2010.
\newblock \emph{Computational Topology: An Introduction}.
\newblock American Mathematical Society.
\newblock ISBN 978-0-8218-4925-5.

\bibitem[{Fefferman, Mitter, and Narayanan(2016)}]{fefferman2016testing}
Fefferman, C.; Mitter, S.; and Narayanan, H. 2016.
\newblock Testing the manifold hypothesis.
\newblock \emph{Journal of the American Mathematical Society}, 29(4):
  983--1049.

\bibitem[{Foroutan et~al.(2024)Foroutan, Rebain, Yi, and
  Tagliasacchi}]{foroutan2024does}
Foroutan, Y.; Rebain, D.; Yi, K.~M.; and Tagliasacchi, A. 2024.
\newblock Does Gaussian Splatting need SFM Initialization?
\newblock \emph{arXiv preprint arXiv:2404.12547}.

\bibitem[{Fridovich-Keil et~al.(2022)Fridovich-Keil, Yu, Tancik, Chen, Recht,
  and Kanazawa}]{fridovich2022plenoxels}
Fridovich-Keil, S.; Yu, A.; Tancik, M.; Chen, Q.; Recht, B.; and Kanazawa, A.
  2022.
\newblock Plenoxels: Radiance fields without neural networks.
\newblock In \emph{Proceedings of the IEEE/CVF Conference on Computer Vision
  and Pattern Recognition}, 5501--5510.

\bibitem[{Gabrielsson et~al.(2020)Gabrielsson, Nelson, Dwaraknath, and
  Skraba}]{gabrielsson2020topology}
Gabrielsson, R.~B.; Nelson, B.~J.; Dwaraknath, A.; and Skraba, P. 2020.
\newblock A topology layer for machine learning.
\newblock In \emph{International Conference on Artificial Intelligence and
  Statistics}, 1553--1563. PMLR.

\bibitem[{Gu{\'e}don and Lepetit(2024)}]{guedon2024sugar}
Gu{\'e}don, A.; and Lepetit, V. 2024.
\newblock Sugar: Surface-aligned gaussian splatting for efficient 3d mesh
  reconstruction and high-quality mesh rendering.
\newblock In \emph{Proceedings of the IEEE/CVF Conference on Computer Vision
  and Pattern Recognition}, 5354--5363.

\bibitem[{Hatcher(2000)}]{Hatcher:478079}
Hatcher, A. 2000.
\newblock \emph{{Algebraic topology}}.
\newblock Cambridge: Cambridge Univ. Press.

\bibitem[{Hedman et~al.(2018)Hedman, Philip, Price, Frahm, Drettakis, and
  Brostow}]{hedman2018deep}
Hedman, P.; Philip, J.; Price, T.; Frahm, J.-M.; Drettakis, G.; and Brostow, G.
  2018.
\newblock Deep blending for free-viewpoint image-based rendering.
\newblock \emph{ACM Transactions on Graphics (ToG)}, 37(6): 1--15.

\bibitem[{Hu et~al.(2024)Hu, Fei, Xu, Hou, Yang, Wang, Lei, Qian, and
  He}]{hu2024topology}
Hu, J.; Fei, B.; Xu, B.; Hou, F.; Yang, W.; Wang, S.; Lei, N.; Qian, C.; and
  He, Y. 2024.
\newblock Topology-Aware Latent Diffusion for 3D Shape Generation.
\newblock \emph{arXiv preprint arXiv:2401.17603}.

\bibitem[{Hu et~al.(2019)Hu, Li, Samaras, and Chen}]{hu2019topology}
Hu, X.; Li, F.; Samaras, D.; and Chen, C. 2019.
\newblock Topology-preserving deep image segmentation.
\newblock \emph{Advances in neural information processing systems}, 32.

\bibitem[{Huang et~al.(2024)Huang, Yu, Chen, Geiger, and Gao}]{huang20242d}
Huang, B.; Yu, Z.; Chen, A.; Geiger, A.; and Gao, S. 2024.
\newblock 2d gaussian splatting for geometrically accurate radiance fields.
\newblock In \emph{ACM SIGGRAPH 2024 Conference Papers}, 1--11.

\bibitem[{Huber(2021)}]{huber2021persistent}
Huber, S. 2021.
\newblock Persistent homology in data science.
\newblock In \emph{Data Science--Analytics and Applications: Proceedings of the
  3rd International Data Science Conference--iDSC2020}, 81--88. Springer.

\bibitem[{Jignasu et~al.(2024)Jignasu, Balu, Sarkar, Hegde,
  Ganapathysubramanian, and Krishnamurthy}]{jignasu2024sdfconnect}
Jignasu, A.; Balu, A.; Sarkar, S.; Hegde, C.; Ganapathysubramanian, B.; and
  Krishnamurthy, A. 2024.
\newblock SDFConnect: Neural Implicit Surface Reconstruction of a Sparse Point
  Cloud with Topological Constraints.
\newblock In \emph{Proceedings of the IEEE/CVF Conference on Computer Vision
  and Pattern Recognition}, 5271--5279.

\bibitem[{Jin et~al.(2021{\natexlab{a}})Jin, Netrapalli, Ge, Kakade, and
  Jordan}]{jin2021nonconvex}
Jin, C.; Netrapalli, P.; Ge, R.; Kakade, S.~M.; and Jordan, M.~I.
  2021{\natexlab{a}}.
\newblock On nonconvex optimization for machine learning: Gradients,
  stochasticity, and saddle points.
\newblock \emph{Journal of the ACM (JACM)}, 68(2): 1--29.

\bibitem[{Jin et~al.(2021{\natexlab{b}})Jin, Mishkin, Mishchuk, Matas, Fua, Yi,
  and Trulls}]{jin2021image}
Jin, Y.; Mishkin, D.; Mishchuk, A.; Matas, J.; Fua, P.; Yi, K.~M.; and Trulls,
  E. 2021{\natexlab{b}}.
\newblock Image matching across wide baselines: From paper to practice.
\newblock \emph{International Journal of Computer Vision}, 129(2): 517--547.

\bibitem[{Kerber, Morozov, and Nigmetov(2017)}]{kerber2017geometry}
Kerber, M.; Morozov, D.; and Nigmetov, A. 2017.
\newblock Geometry helps to compare persistence diagrams.

\bibitem[{Kerbl et~al.(2023)Kerbl, Kopanas, Leimk{\"u}hler, and
  Drettakis}]{kerbl3Dgaussians}
Kerbl, B.; Kopanas, G.; Leimk{\"u}hler, T.; and Drettakis, G. 2023.
\newblock 3D Gaussian Splatting for Real-Time Radiance Field Rendering.
\newblock \emph{ACM Transactions on Graphics}, 42(4).

\bibitem[{Knapitsch et~al.(2017)Knapitsch, Park, Zhou, and
  Koltun}]{knapitsch2017tanks}
Knapitsch, A.; Park, J.; Zhou, Q.-Y.; and Koltun, V. 2017.
\newblock Tanks and temples: Benchmarking large-scale scene reconstruction.
\newblock \emph{ACM Transactions on Graphics (ToG)}, 36(4): 1--13.

\bibitem[{Lee et~al.(2025)Lee, Lee, Sun, Ali, and Park}]{lee2025deblurring}
Lee, B.; Lee, H.; Sun, X.; Ali, U.; and Park, E. 2025.
\newblock Deblurring 3d gaussian splatting.
\newblock In \emph{European Conference on Computer Vision}, 127--143. Springer.

\bibitem[{Li et~al.(2024)Li, Zhang, Bai, Zheng, Ning, Zhou, and
  Gu}]{li2024dngaussian}
Li, J.; Zhang, J.; Bai, X.; Zheng, J.; Ning, X.; Zhou, J.; and Gu, L. 2024.
\newblock Dngaussian: Optimizing sparse-view 3d gaussian radiance fields with
  global-local depth normalization.
\newblock In \emph{Proceedings of the IEEE/CVF Conference on Computer Vision
  and Pattern Recognition}, 20775--20785.

\bibitem[{Lu et~al.(2024)Lu, Yu, Xu, Xiangli, Wang, Lin, and
  Dai}]{lu2024scaffold}
Lu, T.; Yu, M.; Xu, L.; Xiangli, Y.; Wang, L.; Lin, D.; and Dai, B. 2024.
\newblock Scaffold-gs: Structured 3d gaussians for view-adaptive rendering.
\newblock In \emph{Proceedings of the IEEE/CVF Conference on Computer Vision
  and Pattern Recognition}, 20654--20664.

\bibitem[{Maria et~al.(2014)Maria, Boissonnat, Glisse, and
  Yvinec}]{maria2014gudhi}
Maria, C.; Boissonnat, J.-D.; Glisse, M.; and Yvinec, M. 2014.
\newblock The gudhi library: Simplicial complexes and persistent homology.
\newblock In \emph{Mathematical Software--ICMS 2014: 4th International
  Congress, Seoul, South Korea, August 5-9, 2014. Proceedings 4}, 167--174.
  Springer.

\bibitem[{Melodia and Lenz(2020)}]{melodia2020persistent}
Melodia, L.; and Lenz, R. 2020.
\newblock Persistent homology as stopping-criterion for voronoi interpolation.
\newblock In \emph{International Workshop on Combinatorial Image Analysis},
  29--44. Springer.

\bibitem[{Mildenhall et~al.(2021)Mildenhall, Srinivasan, Tancik, Barron,
  Ramamoorthi, and Ng}]{mildenhall2021nerf}
Mildenhall, B.; Srinivasan, P.~P.; Tancik, M.; Barron, J.~T.; Ramamoorthi, R.;
  and Ng, R. 2021.
\newblock Nerf: Representing scenes as neural radiance fields for view
  synthesis.
\newblock \emph{Communications of the ACM}, 65(1): 99--106.

\bibitem[{Muhammad and Yeasin(2020)}]{muhammad2020eigen}
Muhammad, M.~B.; and Yeasin, M. 2020.
\newblock Eigen-cam: Class activation map using principal components.
\newblock In \emph{2020 International Joint Conference on Neural Networks
  (IJCNN)}, 1--7. IEEE.

\bibitem[{M{\"u}ller et~al.(2022)M{\"u}ller, Evans, Schied, and
  Keller}]{muller2022instant}
M{\"u}ller, T.; Evans, A.; Schied, C.; and Keller, A. 2022.
\newblock Instant neural graphics primitives with a multiresolution hash
  encoding.
\newblock \emph{ACM transactions on graphics (TOG)}, 41(4): 1--15.

\bibitem[{Munkres(1984)}]{Munkers84}
Munkres, J.~R. 1984.
\newblock \emph{{Elements of Algebraic Topology}}.
\newblock Addison Wesley Publishing Company.
\newblock ISBN 0201045869.

\bibitem[{Nesterov(2013)}]{nesterov2013introductory}
Nesterov, Y. 2013.
\newblock \emph{Introductory lectures on convex optimization: A basic course},
  volume~87.
\newblock Springer Science \& Business Media.

\bibitem[{Niemeyer et~al.(2024)Niemeyer, Manhardt, Rakotosaona, Oechsle,
  Duckworth, Gosula, Tateno, Bates, Kaeser, and Tombari}]{niemeyer2024radsplat}
Niemeyer, M.; Manhardt, F.; Rakotosaona, M.-J.; Oechsle, M.; Duckworth, D.;
  Gosula, R.; Tateno, K.; Bates, J.; Kaeser, D.; and Tombari, F. 2024.
\newblock Radsplat: Radiance field-informed gaussian splatting for robust
  real-time rendering with 900+ fps.
\newblock \emph{arXiv preprint arXiv:2403.13806}.

\bibitem[{Nishikawa, Ike, and Yamanishi(2024)}]{nishikawa2024adaptive}
Nishikawa, N.; Ike, Y.; and Yamanishi, K. 2024.
\newblock Adaptive Topological Feature via Persistent Homology: Filtration
  Learning for Point Clouds.
\newblock \emph{Advances in Neural Information Processing Systems}, 36.

\bibitem[{Preparata and Shamos(2012)}]{preparata2012computational}
Preparata, F.~P.; and Shamos, M.~I. 2012.
\newblock \emph{Computational geometry: an introduction}.
\newblock Springer Science \& Business Media.

\bibitem[{Pun, Lee, and Xia(2022)}]{pun2022persistent}
Pun, C.~S.; Lee, S.~X.; and Xia, K. 2022.
\newblock Persistent-homology-based machine learning: a survey and a
  comparative study.
\newblock \emph{Artificial Intelligence Review}, 55(7): 5169--5213.

\bibitem[{Qi et~al.(2023)Qi, He, Qi, Zhang, and Yang}]{qi2023dynamic}
Qi, Y.; He, Y.; Qi, X.; Zhang, Y.; and Yang, G. 2023.
\newblock Dynamic snake convolution based on topological geometric constraints
  for tubular structure segmentation.
\newblock In \emph{Proceedings of the IEEE/CVF International Conference on
  Computer Vision}, 6070--6079.

\bibitem[{Reinauer, Caorsi, and Berkouk(2021)}]{reinauer2021persformer}
Reinauer, R.; Caorsi, M.; and Berkouk, N. 2021.
\newblock Persformer: A transformer architecture for topological machine
  learning.
\newblock \emph{arXiv preprint arXiv:2112.15210}.

\bibitem[{Rota~Bul{\`o}, Porzi, and Kontschieder(2025)}]{rota2025revising}
Rota~Bul{\`o}, S.; Porzi, L.; and Kontschieder, P. 2025.
\newblock Revising Densification in Gaussian Splatting.
\newblock In \emph{European Conference on Computer Vision}, 347--362. Springer.

\bibitem[{Sch\"{o}nberger and Frahm(2016)}]{schoenberger2016sfm}
Sch\"{o}nberger, J.~L.; and Frahm, J.-M. 2016.
\newblock Structure-from-Motion Revisited.
\newblock In \emph{Conference on Computer Vision and Pattern Recognition
  (CVPR)}.

\bibitem[{Simonyan(2014)}]{simonyan2014very}
Simonyan, K. 2014.
\newblock Very deep convolutional networks for large-scale image recognition.
\newblock \emph{arXiv preprint arXiv:1409.1556}.

\bibitem[{Skraba and Turner(2020)}]{Skraba2020WassersteinSF}
Skraba, P.; and Turner, K. 2020.
\newblock Wasserstein Stability for Persistence Diagrams.
\newblock \emph{arXiv: Algebraic Topology}.

\bibitem[{Song et~al.(2024)Song, Zheng, Yuan, Gao, Zhao, He, Gu, and
  Zhao}]{song2024sa}
Song, X.; Zheng, J.; Yuan, S.; Gao, H.-a.; Zhao, J.; He, X.; Gu, W.; and Zhao,
  H. 2024.
\newblock SA-GS: Scale-Adaptive Gaussian Splatting for Training-Free
  Anti-Aliasing.
\newblock \emph{arXiv preprint arXiv:2403.19615}.

\bibitem[{Tanemura, Ogawa, and Ogita(1983)}]{tanemura1983new}
Tanemura, M.; Ogawa, T.; and Ogita, N. 1983.
\newblock A new algorithm for three-dimensional Voronoi tessellation.
\newblock \emph{Journal of Computational Physics}, 51(2): 191--207.

\bibitem[{Turkulainen et~al.(2024)Turkulainen, Ren, Melekhov, Seiskari, Rahtu,
  and Kannala}]{turkulainen2024dn}
Turkulainen, M.; Ren, X.; Melekhov, I.; Seiskari, O.; Rahtu, E.; and Kannala,
  J. 2024.
\newblock DN-Splatter: Depth and Normal Priors for Gaussian Splatting and
  Meshing.
\newblock \emph{arXiv preprint arXiv:2403.17822}.

\bibitem[{Ververas et~al.(2025)Ververas, Potamias, Song, Deng, and
  Zafeiriou}]{ververas2025sags}
Ververas, E.; Potamias, R.~A.; Song, J.; Deng, J.; and Zafeiriou, S. 2025.
\newblock Sags: Structure-aware 3d gaussian splatting.
\newblock In \emph{European Conference on Computer Vision}, 221--238. Springer.

\bibitem[{Wang et~al.(2020)Wang, Liu, Samaras, and Chen}]{wang2020topogan}
Wang, F.; Liu, H.; Samaras, D.; and Chen, C. 2020.
\newblock Topogan: A topology-aware generative adversarial network.
\newblock In \emph{Computer Vision--ECCV 2020: 16th European Conference,
  Glasgow, UK, August 23--28, 2020, Proceedings, Part III 16}, 118--136.
  Springer.

\bibitem[{Wang et~al.(2004)Wang, Bovik, Sheikh, and Simoncelli}]{wang2004image}
Wang, Z.; Bovik, A.~C.; Sheikh, H.~R.; and Simoncelli, E.~P. 2004.
\newblock Image quality assessment: from error visibility to structural
  similarity.
\newblock \emph{IEEE transactions on image processing}, 13(4): 600--612.

\bibitem[{Wu et~al.(2024)Wu, Yuan, Zhang, Yang, Cao, Yan, and
  Gao}]{wu2024recent}
Wu, T.; Yuan, Y.-J.; Zhang, L.-X.; Yang, J.; Cao, Y.-P.; Yan, L.-Q.; and Gao,
  L. 2024.
\newblock Recent advances in 3d gaussian splatting.
\newblock \emph{Computational Visual Media}, 1--30.

\bibitem[{Xiangli et~al.(2022)Xiangli, Xu, Pan, Zhao, Rao, Theobalt, Dai, and
  Lin}]{xiangli2022bungeenerf}
Xiangli, Y.; Xu, L.; Pan, X.; Zhao, N.; Rao, A.; Theobalt, C.; Dai, B.; and
  Lin, D. 2022.
\newblock Bungeenerf: Progressive neural radiance field for extreme multi-scale
  scene rendering.
\newblock In \emph{European conference on computer vision}, 106--122. Springer.

\bibitem[{Xu et~al.(2023)Xu, Xiangli, Peng, Pan, Zhao, Theobalt, Dai, and
  Lin}]{xu2023grid}
Xu, L.; Xiangli, Y.; Peng, S.; Pan, X.; Zhao, N.; Theobalt, C.; Dai, B.; and
  Lin, D. 2023.
\newblock Grid-guided neural radiance fields for large urban scenes.
\newblock In \emph{Proceedings of the IEEE/CVF Conference on Computer Vision
  and Pattern Recognition}, 8296--8306.

\bibitem[{Yamada and Shibuya(2020)}]{yamada2020inferring}
Yamada, N.; and Shibuya, T. 2020.
\newblock Inferring Underlying Manifold of Low Density Data using Adaptive
  Interpolation.
\newblock In \emph{ICAART (2)}, 395--402.

\bibitem[{Yan et~al.(2021)Yan, Hu, Mao, Ye, and Yu}]{yan2021deep}
Yan, X.; Hu, S.; Mao, Y.; Ye, Y.; and Yu, H. 2021.
\newblock Deep multi-view learning methods: A review.
\newblock \emph{Neurocomputing}, 448: 106--129.

\bibitem[{Yang et~al.(2024{\natexlab{a}})Yang, Zhu, Jiang, Ye, Chen, Zhang,
  Chen, Zhao, and Zhao}]{yang2024spectrally}
Yang, R.; Zhu, Z.; Jiang, Z.; Ye, B.; Chen, X.; Zhang, Y.; Chen, Y.; Zhao, J.;
  and Zhao, H. 2024{\natexlab{a}}.
\newblock Spectrally Pruned Gaussian Fields with Neural Compensation.
\newblock \emph{arXiv preprint arXiv:2405.00676}.

\bibitem[{Yang et~al.(2024{\natexlab{b}})Yang, Gao, Sun, Huang, Lyu, Zhou,
  Jiao, Qi, and Jin}]{yang2024spec}
Yang, Z.; Gao, X.; Sun, Y.; Huang, Y.; Lyu, X.; Zhou, W.; Jiao, S.; Qi, X.; and
  Jin, X. 2024{\natexlab{b}}.
\newblock Spec-gaussian: Anisotropic view-dependent appearance for 3d gaussian
  splatting.
\newblock \emph{arXiv preprint arXiv:2402.15870}.

\bibitem[{Yu et~al.(2024)Yu, Chen, Huang, Sattler, and Geiger}]{yu2024mip}
Yu, Z.; Chen, A.; Huang, B.; Sattler, T.; and Geiger, A. 2024.
\newblock Mip-splatting: Alias-free 3d gaussian splatting.
\newblock In \emph{Proceedings of the IEEE/CVF Conference on Computer Vision
  and Pattern Recognition}, 19447--19456.

\bibitem[{Yu, Sattler, and Geiger(2024)}]{yu2024gaussian}
Yu, Z.; Sattler, T.; and Geiger, A. 2024.
\newblock Gaussian opacity fields: Efficient adaptive surface reconstruction in
  unbounded scenes.
\newblock \emph{ACM Transactions on Graphics (TOG)}, 43(6): 1--13.

\bibitem[{Yuan et~al.(2025{\natexlab{a}})Yuan, Luo, Shen, Li, Liu, Mao, and
  Wang}]{yuan2025dvp}
Yuan, Z.; Luo, J.; Shen, F.; Li, Z.; Liu, C.; Mao, T.; and Wang, Z.
  2025{\natexlab{a}}.
\newblock DVP-MVS: Synergize depth-edge and visibility prior for multi-view
  stereo.
\newblock In \emph{Proceedings of the AAAI Conference on Artificial
  Intelligence}, volume~39, 9743--9752.

\bibitem[{Yuan et~al.(2025{\natexlab{b}})Yuan, Yang, Cai, Wu, Liu, Zhang,
  Jiang, Li, and Wang}]{yuan2025sed}
Yuan, Z.; Yang, Z.; Cai, Y.; Wu, K.; Liu, M.; Zhang, D.; Jiang, H.; Li, Z.; and
  Wang, Z. 2025{\natexlab{b}}.
\newblock SED-MVS: Segmentation-Driven and Edge-Aligned Deformation Multi-View
  Stereo with Depth Restoration and Occlusion Constraint.
\newblock \emph{IEEE Transactions on Circuits and Systems for Video
  Technology}.

\bibitem[{Zhang et~al.(2020)Zhang, Riegler, Snavely, and
  Koltun}]{zhang2020nerf++}
Zhang, K.; Riegler, G.; Snavely, N.; and Koltun, V. 2020.
\newblock Nerf++: Analyzing and improving neural radiance fields.
\newblock \emph{arXiv preprint arXiv:2010.07492}.

\bibitem[{Zhang et~al.(2018)Zhang, Isola, Efros, Shechtman, and
  Wang}]{zhang2018unreasonable}
Zhang, R.; Isola, P.; Efros, A.~A.; Shechtman, E.; and Wang, O. 2018.
\newblock The unreasonable effectiveness of deep features as a perceptual
  metric.
\newblock In \emph{Proceedings of the IEEE conference on computer vision and
  pattern recognition}, 586--595.

\bibitem[{Zhang et~al.(2022)Zhang, Yao, Wang, and Chen}]{zhang2022convergence}
Zhang, Y.; Yao, J.; Wang, Y.; and Chen, C. 2022.
\newblock On the convergence of optimizing persistent-homology-based losses.
\newblock \emph{arXiv preprint arXiv:2206.02946}.

\bibitem[{Zhang et~al.(2025)Zhang, Hu, Lao, He, and Zhao}]{zhang2025pixelgs}
Zhang, Z.; Hu, W.; Lao, Y.; He, T.; and Zhao, H. 2025.
\newblock Pixel-GS: Density Control with Pixel-Aware Gradient for 3D Gaussian
  Splatting.
\newblock In \emph{European Conference on Computer Vision}, 326--342. Springer.

\bibitem[{Zhu et~al.(2025)Zhu, Fan, Jiang, and Wang}]{zhu2025fsgs}
Zhu, Z.; Fan, Z.; Jiang, Y.; and Wang, Z. 2025.
\newblock Fsgs: Real-time few-shot view synthesis using gaussian splatting.
\newblock In \emph{European Conference on Computer Vision}, 145--163. Springer.

\bibitem[{Zia et~al.(2024)Zia, Khamis, Nichols, Tayab, Hayder, Rolland, Stone,
  and Petersson}]{zia2024topological}
Zia, A.; Khamis, A.; Nichols, J.; Tayab, U.~B.; Hayder, Z.; Rolland, V.; Stone,
  E.; and Petersson, L. 2024.
\newblock Topological deep learning: A review of an emerging paradigm.
\newblock \emph{Artificial Intelligence Review}, 57(4): 77.

\bibitem[{Zou et~al.(2024)Zou, Yu, Guo, Li, Liang, Cao, and
  Zhang}]{zou2024triplane}
Zou, Z.-X.; Yu, Z.; Guo, Y.-C.; Li, Y.; Liang, D.; Cao, Y.-P.; and Zhang, S.-H.
  2024.
\newblock Triplane meets gaussian splatting: Fast and generalizable single-view
  3d reconstruction with transformers.
\newblock In \emph{Proceedings of the IEEE/CVF Conference on Computer Vision
  and Pattern Recognition}, 10324--10335.

\end{thebibliography}

\section{Appendices*}
\subsection{A. REFORMULATION OF NOTATIONS}
In order to maintain clarity, we adopt a notation that aligns with established conventions in the field. This allows us to effectively reformulate the concept related to this paper.
\subsubsection{A1. Reformulation of Persistent Homology}
Persistent homology is a mathematical framework for elucidating topological features in data across multiple scales, with provable robustness guarantees. By leveraging the formalism of algebraic topology \cite{Hatcher:478079, Munkers84}, persistent homology enables the systematic analysis of connected components, loops, voids, and higher-dimensional analogues. Specifically, given a domain $X$, a filter function $f : X \to \mathbb{R}$ is employed to capture and quantify these topological structures.
We employ a filter function to filter the domain, incrementally increasing the threshold value $\alpha$. As $\alpha$ grows, the sublevel set $X_\alpha = \{ x \in X \mid f(x) \leq \alpha \}$ expands continuously. This process yields a sequence of nested spaces, known as a filtration of $X$: $\emptyset = X_{-\infty} \subset \ldots \subset X_{\infty} = X$. Throughout this filtration, topological structures emerge (birth) and vanish (death) as $X_\alpha$ evolves from $\emptyset$ to $X$. This process is illustrated in Figure 3(a-d) of the main text, where the space is filtered using a distance transform derived from the input data.
By applying the homology functor to the filtration, we obtain a precise quantification of the birth and death of topological features, as captured by homology groups, resulting in the persistence diagram (PD). A PD is a planar multiset of points, each representing a homological feature, such as components, loops, and their higher-dimensional analogs. Each point $p = (b, d)$ corresponds to the birth and death times of a feature during the filtration, with its lifetime, $\text{Pers}(p) = |d - b|$, also known as its lifespan in the main text, but referred to as persistence here, measuring its importance with respect to the input filter function. We augment the diagram with the diagonal line $\Delta$ for technical reasons. We denote the PD of a filter function $f$ by $\text{Dgm}(f)$.

\subsubsection{A2. Reformulation of Simplicial Complex}
In computational settings, we often discretize the domain into a simplicial complex, comprising a set of simplices such as vertices, edges, and triangles. Assuming the complex remains static throughout optimization, we assign filter function values to vertices and extend them to all simplices. Specifically, a simplex's filter function value is the maximum of its vertices, i.e., $f(\sigma) = \max_{v \in \sigma} f(v)$. We then construct the filtration as a sequence of subcomplexes, which enables the computation of persistent homology via a reduction algorithm on the boundary matrix. This algorithm encodes the combinatorial relationships between elements of the complex.
Each persistent point corresponds to a pair of simplices, with its birth and death times determined by the function values of these simplices, ultimately the function values of two vertices. Formally, for a persistent point $p \in \text{Dgm}(f)$, its birth and death times are $\text{birth}(p) = f(v_b(p))$ and $\text{death}(p) = f(v_d(p))$, respectively, where $v_b(p)$ and $v_d(p)$ are the corresponding birth and death vertices. The gradient of the loss function manipulates these persistent points by adjusting the function values of the relevant vertices, $v_b(p)$ and $v_d(p)$, which is essential for topology-based loss optimization. More technical details can be found in \cite{book111, book222}.

\begin{figure*}[t]
	\centering
	\includegraphics[width=13.2cm]{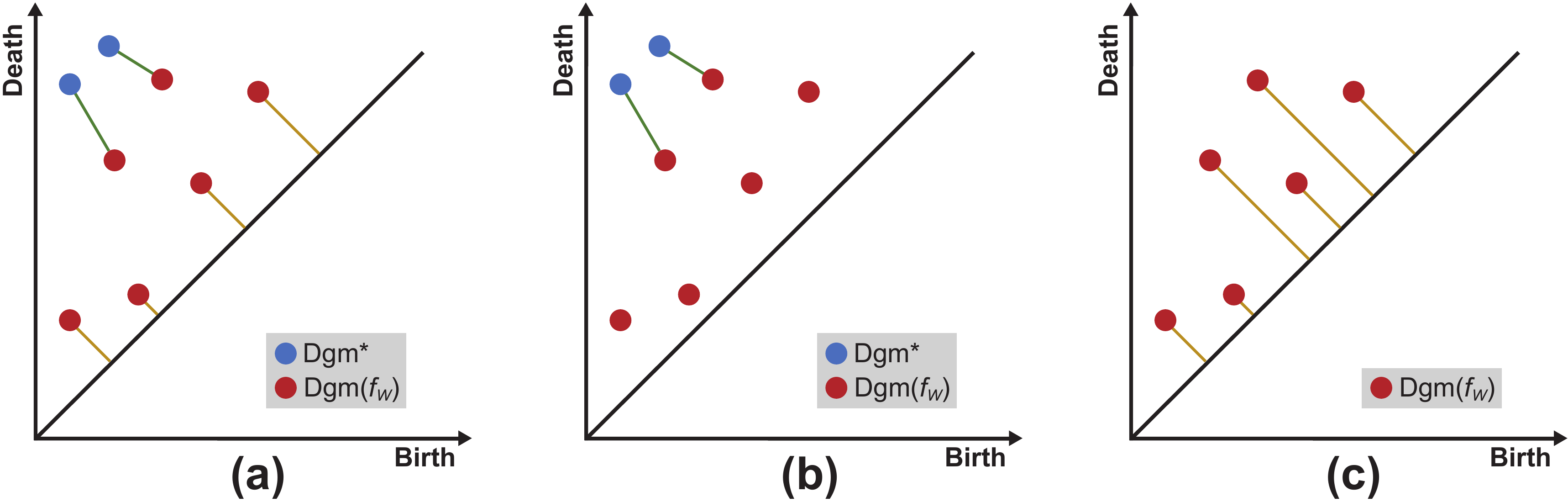}
	\caption{\textbf{a}: Wasserstein distance between two diagrams, the ground truth diagram $\text{Dgm}^*$ and the diagram derived from the model prediction $\text{Dgm}(f_W)$; \textbf{b}: the matching used for topology-based loss term $L_{\text{topo}}$ in \cite{zhang2022convergence}. All matchings between a non-diagonal point of $\text{Dgm}(f_W)$ and the diagonal from the Wasserstein distance (orange lines in panel a) are ignored; \textbf{c}: the total persistence is the cost of matching all non-diagonal points to the diagonal.}
	\label{fig_pd}
\end{figure*}

\subsubsection{A3. Reformulation of Loss Function}
We consider the model's prediction, $\phi_W(x)$, parameterized by learnable weights $W$, and compare it with the ground truth $y$ for every datum $(x, y) \in \mathcal{D}$. The prediction $\phi_W$ induces a filter function $f_W$, which in turn yields a persistence diagram (PD) $\text{Dgm}(f_W)$ that describes the topological properties of the prediction. To evaluate the quality of the prediction, we compare this diagram with a ground truth diagram $\text{Dgm}^*$, which represents the desired topology. Given a specific ground truth function, we can compute $\text{Dgm}^*$. As we are applying persistence homology to images rendered by 3D-GS, which have a finite resolution, the number of true structures $\beta$ is necessarily finite. Therefore, we assume that $\beta$ is upper-bounded by a small number $B$, which is crucial for establishing our theoretical bounds.
In general, a topology-aware total loss function comprises two primary components: a standard supervision loss term $L_{\text{supv}}$ and a topology-based loss term $L_{\text{topo}}$. In the context of this work, the supervision loss term $L_{\text{supv}}$ is specifically defined as a combination of L1 loss and SSIM loss, which are commonly employed in 3D-GS. The topology-based loss term $L_{\text{topo}}$ serves to quantify the similarity between the topological features of the predicted output and the desired topology.
As noted in \cite{zhang2022convergence}, an additional regularization term $L_{\text{reg}}$ can be introduced to ensure satisfactory convergence guarantees, which will be elaborated upon in Subsubsection B2.
In this study, our topology-aware total loss is composed of $L_{\text{supv}}$ and a PersLoss regularization term. The relationship between PersLoss, $L_{\text{topo}}$, and $L_{\text{reg}}$ will be thoroughly discussed in Subsubsection B3.

\subsection{B. DISCUSSION OF DISTANCE AND LOSS}
This section is devoted to a discussion on the distance metric between persistence diagrams and the corresponding loss functions.

\subsubsection{B1. Stability Results of PD Distance}
Persistent homology exhibits stability, a property that guarantees the distance between two persistence diagrams is bounded by the difference between their input functions. Formally, given two tame functions $f, g : X \to \mathbb{R}$, let $\Pi$ denote the set of all bijections between their persistence diagrams. The $q$-Wasserstein distance between $\text{Dgm}(f)$ and $\text{Dgm}(g)$ is defined as (the Wasserstein distance is a well-established metric for measuring the distance between two persistent diagrams):
\begin{equation}
	\inf_{\pi \in \Pi} \left[ \sum_{p \in \text{Dgm}(f)} \left( \text{B}_{\pi}(p)^q + \text{D}_{\pi}(p)^q \right) \right]^{1 / q},
\end{equation}
where,
\begin{equation}
\left\{
	\begin{array}{l}
		\text{B}_{\pi}(p) = \left| \text{birth}(p) - \text{birth}(\pi(p)) \right| \\
		\text{D}_{\pi}(p) = \left| \text{death}(p) - \text{death}(\pi(p)) \right|
	\end{array}.
\right.
\end{equation}

Here, $\text{birth}(p)$ and $\text{death}(p)$ denote the coordinates of the persistence point $p$. The bottleneck distance, a precursor to the Wasserstein distance, is equivalent to the Wasserstein distance with $q = \infty$ \cite{cohen2007stability}.

The Wasserstein Stability Theorem \cite{cohen2010lipschitz, Skraba2020WassersteinSF} asserts that for two tame Lipschitz functions $f$ and $g$ defined on a triangulable compact metric space $X$, there exist constants $k$ and $C$, dependent on $X$ and the Lipschitz constants of $f$ and $g$, such that for all $q \geq k$,
\begin{equation}
	d_q(\text{Dgm}(f), \text{Dgm}(g)) \leq C \cdot \| f - g \|_{\infty}^{1 - k / q}.
\end{equation}

This theorem provides a theoretical foundation for the convergence of the loss function, which will be leveraged in our subsequent analysis, although a detailed discussion is beyond the scope of this paper.

\subsubsection{B2. Regularization Term $L_{\text{reg}}$}
Traditional topology-based loss terms rely on the Wasserstein distance between the persistence diagrams (PDs) of the model output and the ground truth. To enhance optimization guarantees without compromising efficacy, \cite{zhang2022convergence} introduces a regularization term $L_{\text{reg}}$ utilizing the total persistence ($\text{TotPers}$) \cite{cohen2010lipschitz} of the model output PD, $\text{Dgm}(f_W)$. Formally, the $k$-th total persistence of $\text{Dgm}(f_W)$ is defined as:
\begin{align}
	\text{TotPers}_k(f_W) & = \sum_{p \in \text{Dgm}(f_W)} \text{Pers}(p)^k 
	\nonumber \\ & = \sum_{p \in \text{Dgm}(f_W)} \left( \text{death}(p) - \text{birth}(p) \right)^k.
\end{align}

As depicted in Figure \ref{fig_pd}.c, the total persistence can be viewed as the matching distance between the persistence diagram and an empty diagram, where all non-diagonal points are matched to the diagonal of the empty diagram. This measure aggregates the persistence of all points in the diagram, effectively quantifying its "norm". By optimizing this loss term, all diagram points are pulled towards the diagonal, resulting in the "shrinking" of corresponding topological structures. This plays a crucial role in stabilizing the loss function, which is essential for achieving the convergence bound \cite{zhang2022convergence}.

\subsubsection{B3. Relationship between $L_{\text{topo}}$, $L_{\text{reg}}$, and $\text{PersLoss}$}
As illustrated in Figure \ref{fig_pd}.a, traditional topology-based loss terms $L_{\text{topo}}$ quantify the matching cost between the persistence diagram of the model output, $\text{Dgm}(f_W)$ (red points), and the ground truth diagram, $\text{Dgm}^*$ (blue points).
During optimization, the gradient descent step adjusts the model weights, thereby modifying the output function $f_W$ and consequently altering the persistence diagram. 
The red points will move towards their matches, which can be broadly classified into two categories \cite{zhang2022convergence}.
On one hand, certain red points are matched to non-diagonal blue points (highlighted with green lines), corresponding to structures that we seek to "restore" by increasing their persistence and aligning them with a salient structure in the ground truth. This cost is denoted as the restoration cost.
On the other hand, the remaining red points are matched to the diagonal (highlighted with orange lines). During optimization, these points are moved towards the diagonal, effectively "shrinking" the corresponding structures, which are deemed noise. This cost is referred to as the shrinking cost.

In a departure from traditional approaches, \cite{zhang2022convergence} decomposes the $L_{\text{topo}}$ into restoration cost and shrinking cost, treating them separately. The restoration cost is measured using the Wasserstein distance, while the shrinking cost is handled by the regularization term $L_{\text{reg}}$, designed to mitigate the impact of noise structures.
In contrast to \cite{zhang2022convergence}, our method employs $\text{PersLoss}$ to simultaneously address both restoration cost and shrinking cost. By leveraging the Truncated Persistent Barcode, we retain only the top-$k$ longest barcodes, effectively eliminating noise. This approach translates to the persistent diagram, where points distant from the diagonal are optimized using a distance metric inspired by the Wasserstein distance, albeit without the requirement of optimal matching. Conversely, points proximal to the diagonal are directly pulled towards the diagonal, thereby nullifying the shrinking cost and rendering the $L_{\text{reg}}$ term superfluous.

Therefore, we make the following assumption:
\begin{itemize}
	\item \textbf{Assumpt. 1 (A1):} $L_{\text{reg}}=0$.
\end{itemize}
The validity of this assumption hinges on the premise that the value of $k$ in the top-$k$ longest barcodes is relatively small, corresponding to the condition in the main text where $B=k_0+k_1+k_2$ is sufficiently small. 

\subsubsection{B4. Optimization Behavior of $\text{PersLoss}$}
Analyzing the optimization behavior of $\text{PersLoss}$ poses several challenges. Firstly, each persistent point corresponds to a pair of critical simplices, and this correspondence can change as the model or function is updated. Secondly, the definition of $\text{PersLoss}$ relies on a non-optimal matching between the diagrams, where the top-$k$ longest barcodes are matched in order of their lengths. Finally, the matching itself can change as the model or function is updated, regardless of whether it is optimal or not.

To address these challenges, we first make an assumption to alleviate the second issue:
\begin{itemize}
	\item \textbf{Assumpt. 2 (A2):} ${\text{PersLoss}} \leq d_2(\text{Dgm}(f), \text{Dgm}(g))$.
\end{itemize}
This assumption is reasonable, as the $\text{PersLoss}$ omits the distances between numerous noise points, leaving only a small number of points. When $B=k_0+k_1+k_2$ in the main text is sufficiently small, the error between the length-based matching used in $\text{PersLoss}$ and the optimal matching used in the Wasserstein distance is negligible.

Assumption \textbf{A2} enables us to relax the loss function value to an upper bound, effectively transforming the optimization landscape from a non-optimal matching to an optimal matching. This allows us to analyze the optimization behavior of $\text{PersLoss}$ by studying the topology-based loss founded on the Wasserstein distance, while neglecting the regularization term $L_{\text{reg}}$, which is zero according to Assumption \textbf{A1}. 
Therefore, we formally define our topology-aware total loss as follows:
\paragraph{Definition 1 (Total Loss).} The topology-aware total loss is given by:
\begin{equation}
	G(W) = L_{\text{supv}}(W, D) + \lambda_{\text{topo}} L_{\text{topo}}(W, D),
\end{equation}
where,
\begin{equation}
	L_{\text{supv}}(W) = \sum_{(x,y) \in D} \ell(\phi_W(x), y),
\end{equation}
\begin{equation}
	L_{\text{topo}}(W) = \min_{\gamma \in \Gamma} \sum_{p \in \overline{\text{Dgm}^*}} \left[ \text{B}_{\gamma}(p)^2 + \text{D}_{\gamma}(p)^2 \right].
\end{equation}

Here, the combined loss of L1 and SSIM in the 3D-GS is represented by $\ell(\cdot)$. We also modify the traditional matching process between $\text{Dgm}(f_W)$ and $\text{Dgm}^*$. Specifically, we disregard the diagonal line of $\text{Dgm}^*$ and establish an injective correspondence between its off-diagonal points and $\text{Dgm}(f_W)$, as illustrated in Figure \ref{fig_pd}.b. We denote the true diagram with the diagonal removed as $\overline{\text{Dgm}^*} = \text{Dgm}^* \setminus \Delta$. The set of eligible injective mappings from the true diagram to the prediction diagram is defined as:
\begin{equation}
	\Gamma(\text{Dgm}(f_W)) = \gamma: \overline{\text{Dgm}^*} \rightarrow \text{Dgm}(f_W).
\end{equation}

Here, for any two distinct points $p_1$ and $p_2$, we have $\gamma(p_1) \neq \gamma(p_2)$. Notably, our $\text{PersLoss}$ has been relaxed to the optimal matching cost $L_{\text{topo}}$ achievable by any injection within $\Gamma$.

\subsubsection{B5. Optimization Algorithm and its Fluctuation}
The optimization of the Topology-aware Total Loss can be achieved through a general algorithm \cite{zhang2022convergence}, i.e. algorithm \ref{alg:optim}, which is outlined below.

Here, $G_t$, $L^t_{\text{supv}}$, and $L^t_{\text{topo}}$ in algorithm \ref{alg:optim} are topology-aware total loss function and its components evaluated at time $t$, based on the underlying diagram and the matching $\gamma_t$. The optimall matching $\gamma_t$ for the topology-based loss term is defined as:
\begin{equation}
	\gamma_t = \arg \max_{\gamma \in \Gamma (\text{Dgm}(f_t))} \sum_{p \in \overline{\text{Dgm}^*}} \left[ \text{B}_{\gamma}(p)^2 + \text{D}_{\gamma}(p)^2 \right],
\end{equation}
where $f_t = f_{W_t}$ denotes the function with parameter $W_t$ at time $t$, and $\text{Dgm}(f_t)$ is the corresponding diagram.
\begin{algorithm}
	\caption{Optimizing a Topology-Aware Total Loss}
	\label{alg:optim}
	\begin{algorithmic}[1]
		\Require $\mathcal{D}$, $\text{Dgm}^*$, learning rate $\eta$, convergence criterion $\epsilon$, and weights $\lambda_{\text{topo}}$
		\Ensure Model weight $W$ 
		\State Randomly initialize $W_0$
		\For{$t = 0, 1, 2, \ldots, T$}
		\State Compute $\text{Dgm}(f_{W_t})$
		\State Compute $\gamma_t$
		\State Gradient descent: $W_{t+1} = W_t - \eta \nabla_W G_t(W_t)$
		\If{$G_t(W_{t+1}) - G_t(W_t) \leq \epsilon$}
		\State Break the loop. Algorithm converges.
		\EndIf
		\EndFor
		\State \Return $W_{t+1}$
	\end{algorithmic}
\end{algorithm}

It is worth noting that the underlying configuration may not necessarily align with the input parameter. As a reminder, in a persistence diagram, the coordinates of a persistent point $p$ correspond to the function values of its birth and death vertices. When computing the loss at time $t$, we utilize the birth and death vertices from the diagram at time $t$, irrespective of the input parameter. \\
Consequently, we have:
\begin{equation}
	L^t_{\text{topo}}(W_t) = \sum_{p \in \overline{\text{Dgm}^*}} \left[ \text{B}_{\gamma_t}^{f_t}(p)^2 + \text{D}_{\gamma_t}^{f_t}(p)^2 \right],
\end{equation}
\begin{equation}
	L^t_{\text{topo}}(W_{t+1}) = \sum_{p \in \overline{\text{Dgm}^*}} \left[ \text{B}_{\gamma_t}^{f_{t+1}}(p)^2 + \text{D}_{\gamma_t}^{f_{t+1}}(p)^2 \right],
\end{equation}
\begin{equation}
	L^{t+1}_{\text{topo}}(W_{t+1}) = \sum_{p \in \overline{\text{Dgm}^*}} \left[ \text{B}_{\gamma_{t+1}}^{f_{t+1}}(p)^2 + \text{D}_{\gamma_{t+1}}^{f_{t+1}}(p)^2 \right],
\end{equation}
in which,
\begin{equation}
	\text{B}_{\gamma_t}^{f_t}(p) = \left| \text{birth}(p) - f_t(v_b(\gamma_t(p))) \right|,
\end{equation}
\begin{equation}
	\text{D}_{\gamma_t}^{f_t}(p) = \left| \text{death}(p) - f_t(v_d(\gamma_t(p))) \right|.
\end{equation}

Here, the first two loss terms are evaluated using the same matched point $\gamma_t(p)$ and its birth and death vertices, but with different filter functions $f_t$ and $f_{t+1}$. In contrast, the third loss term is evaluated using the new configuration at $t+1$ and the new function $f_{t+1}$. Analogously, we can define the supervision loss terms $L^t_{\text{supv}}(W_t)$, $L^t_{\text{supv}}(W_{t+1})$, and $L^{t+1}_{\text{supv}}(W_{t+1})$. Notably, $L^t_{\text{supv}}(W_{t+1})$ is equivalent to $L^{t+1}_{\text{supv}}(W_{t+1})$, as it does not involve the configuration update. The total loss function $G_t$ is then given by the sum of these terms: $G_t = L^t_{\text{supv}} + \lambda_{\text{topo}} L^t_{\text{topo}}$.

\begin{figure}[t]
	\centering
	\includegraphics[width=7.8cm]{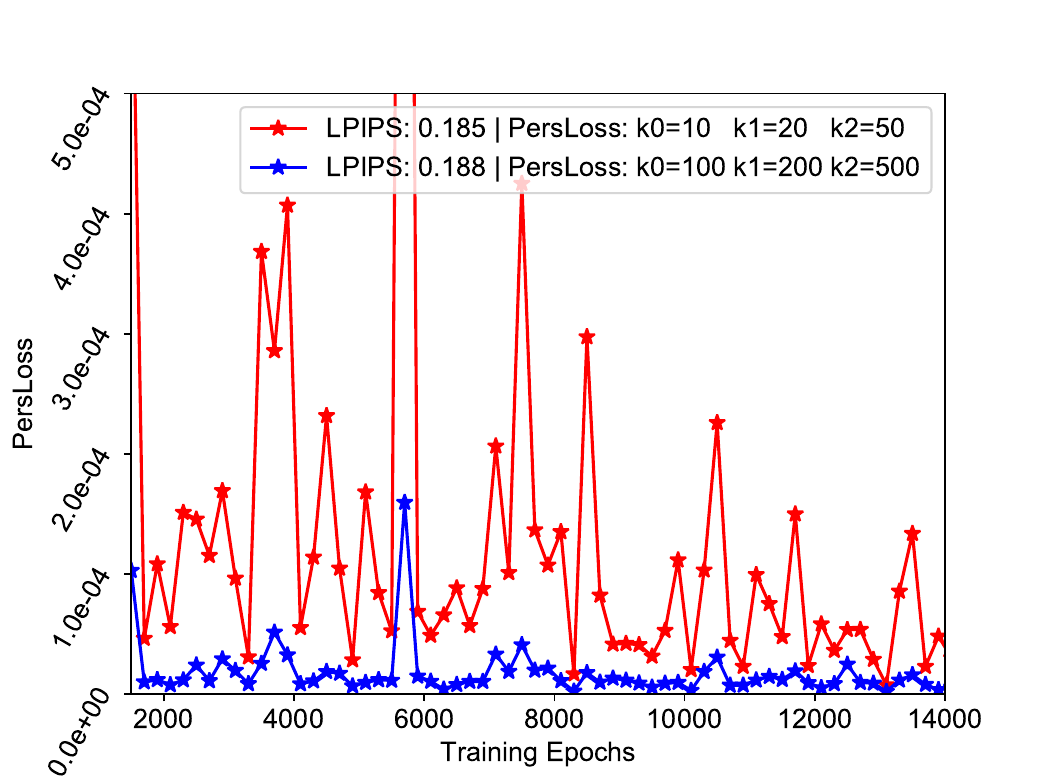}
	\caption{Evolution of the $\text{PersLoss}$ regularization term values (excluding $\lambda_{\text{topo}}$) during the ADC stage on the room scene from \cite{barron2022mip}. The red curve corresponds to the preservation of only significant topological features, whereas the blue curve represents the inclusion of additional noise persistence points.}
	\label{fig_persloss}
\end{figure}

During the optimization process, the loss function is updated as the parameters are updated. Specifically, the update rule is given by:
\begin{equation}
	\cdots G_t(W_t) \rightarrow G_t(W_{t+1}) \rightarrow G_{t+1}(W_{t+1}) \cdots.
\end{equation}
The optimization process consists of two distinct steps. Firstly, we compute the gradient using the configuration at time $t$ and update the parameter to $W_{t+1}$ via gradient descent. This step guarantees a monotonic decrease in the loss function. Secondly, we keep the parameter fixed and update the underlying configuration to reflect the new diagram $\text{Dgm}(f_{t+1})$ and the new optimal match $\gamma_{t+1}$. Notably, this step may result in an increase in the loss function. 

As illustrated in Figure \ref{fig_persloss}, the interplay between the two steps induces fluctuations in the Topology-aware Total Loss (as well as the $\text{PersLoss}$).
When $B$ is small, the retained significant persistence points effectively guide the Gaussians to learn the desired topological structures within a limited number of iterations during the ADC stage. Consequently, $\text{PersLoss}$ exhibits a larger fluctuation amplitude but overall rapidly decreases, converging towards a stable value.
In contrast, if $B$ is large, a substantial number of noise points near the diagonal may be retained. In this scenario, $\text{PersLoss}$ may rapidly optimize these noise points, resulting in an extremely small regularization term value. However, such a small value is insufficient to guide the overall topology-aware total loss, thereby hindering the prediction of topological features that closely approximate the desired topology.

A key theoretical contribution of this work is to demonstrate that the decrease in loss during the first step dominates the potential increase in loss during the second step, thereby addressing the two remaining challenges posed in Subsubsection B4. For a formal statement and proof of our main theorem, we refer the reader to Subsection C.

\subsection{C. THEOREM OF CONVERGENCE RATE}
This section establishes and proves our main result on the time complexity of the optimization algorithm, with the $O(\cdot)$ notation hiding constants that do not depend on $\epsilon$.

\subsubsection{C1. Setup of Main Theoretical}
We begin by introducing the following assumptions, consistent with those in the main text, concerning the behavior of the filter function $f_W$ and the supervision loss function $L_{\text{supv}}(W)$. These assumptions are standard regularity conditions in the analysis of convergence for optimization algorithms and topological data analysis.

\begin{itemize}
	\item \textbf{Assumpt. 3 (A3):} $f$ is 1-bounded, 1-Lipschitz continuous and 1-Lipschitz smooth relative to $W$.
	\item \textbf{Assumpt. 4 (A4):} $L_{\text{supv}}(W)$ is $\ell^0$-bounded, $\ell^1$-Lipschitz continuous and $\ell^2$-Lipschitz smooth relative to $W$.
\end{itemize}

We demonstrate that a judicious choice of step-size/learning rate enables control over the optimization process, ensuring that the increase in topological loss due to persistence diagram updates, $G_{t+1}(W_{t+1}) - G_t(W_{t+1})$, is always offset by the overall loss reduction $G_t(W_t) - G_t(W_{t+1})$. Specifically, this guarantees a monotonic decrease in the total loss function at each iteration, thereby ensuring efficient convergence and termination.

\paragraph{Theorem 1.} \textit{Under Assumptions \textbf{A1}-\textbf{A4}, and a prescribed stopping condition $\epsilon$, we establish that the optimization algorithm utilizing our topology-aware total loss converges in $O\left(\frac{1}{\epsilon}\right)$ iterations, provided that the step-size $\eta$ is selected as:}
\begin{equation}
	\eta \leq \min \left\{
	\frac{1}{2\ell^2 + 10\lambda_{\text{topo}}B}, 
	\frac{\epsilon}{4096\lambda_{\text{topo}}^2 B^2}
	\right\}.
\end{equation}

Here, $B=k_0+k_1+k_2$ is the cardinality of the ground truth diagram (excluding the diagonal), i.e., $B = \text{card}(\overline{\text{Dgm}^*})$.

\subsubsection{C2. Technical Lemmas of Main Theorem}
The proof of Theorem 1 relies on the following lemmas.

\paragraph{Lemma 1.} \textit{Assume \textbf{A3} holds, we have:}
\begin{enumerate}
	\item $\lambda_{\text{topo}}L_{\text{topo}}(W_t) \leq \lambda_{\text{topo}}B$
	\item $\|\nabla_W \lambda_{\text{topo}}L_{\text{topo}}(W_t)\|_2 \leq 2\lambda_{\text{topo}}B$
	\item $\|\nabla_W^2 \lambda_{\text{topo}}L_{\text{topo}}(W_t) \|_2 \leq  5\lambda_{\text{topo}}B$
\end{enumerate}

The first bound is a consequence of \textbf{A3}, while the second bound follows from the fact that the first-order derivative $|\nabla_W \lambda_{\text{topo}}L_{\text{topo}}(W_t)|_2$ is bounded by $2B$. The third bound is obtained by noting that $|\nabla_W^2 \lambda{\text{topo}}L_{\text{topo}}(W_t)|_2$ consists of quadratic functions, which are bounded by $B+4B=5B$.

Combining Lemma 1 and Assumption \textbf{A4}, we can deduce the following bounds on the derivatives of $G_t(W_t)$ up to second order:

\textbf{Fact 1: Bounded function value:} $G_t(W_t) \leq \ell^0 + \lambda_{\text{topo}}B \triangleq C_0$

\textbf{Fact 2: Bounded gradient:} $\|\nabla_W G_t(W_t)\|_2 \leq \ell^1 + 2\lambda_{\text{topo}}B \triangleq C_1$

\textbf{Fact 3: Bounded Hessian:} $\| \nabla_W^2 G_t(W_t)\|_2 \leq \ell^2 + 5\lambda_{\text{topo}}B \triangleq C_2$

Here, we denote these bounds by $C_0$, $C_1$, and $C_2$ for ease of notation. This lemma serves as a foundation for the following result.

\paragraph{Lemma 2.} (Improve or Localize \cite{jin2021nonconvex}). \textit{The magnitude of the parameter update is bounded by the step-size and the change in the topology-aware total loss function.}
	\begin{equation}
		\|W_{t+1} - W_t\| \leq 2\sqrt{\eta}(G_t(W_t) - G_t(W_{t+1})).
	\end{equation}

\textit{Proof.} Following \cite{jin2021nonconvex} and \cite{nesterov2013introductory}, we can write:
\begin{align}
	& G_t(W_{t+1}) \leq G_t(W_t) + 
	\nonumber \\
	& \nabla G_t(W_t)^\top [W_{t+1} - W_t] + \frac{C_2}{2} \|W_{t+1} - W_t\|^2.
\end{align}

Using the update equation $W_{t+1} = W_t - \eta \nabla_W G_t(W_t)$ and selecting $\eta \leq \frac{1}{2C_2}$, we derive the following inequality:
\begin{align}
	G_t(W_t) - G_t(W_{t+1}) & \geq \frac{\eta}{4} \|\nabla_W G_t(W_t)\|_2^2 \nonumber \\
		& = \frac{1}{4\eta} \|W_{t+1} - W_t\|^2.
\end{align}

This inequality directly implies the result of the lemma. $\square$

\paragraph{Lemma 3.} (Bounded Increase of the Topology-based Loss Term) \textit{The increase in $L_{\text{topo}}$ due to the configuration change is bounded as follows:}
\begin{align}
	&L_{\text{topo}}^t(W_{t+1}) - L_{\text{topo}}^{t+1}(W_{t+1}) \nonumber \\
	&\leq 16 B \sqrt{\eta} (G_t(W_t) - G_t(W_{t+1})).
\end{align}

\textit{Proof.} The proof can be found in \cite{zhang2022convergence}. $\square$

\subsubsection{C3. Proof of the Main Theorem}
We are now in a position to prove our main theorem. To begin, we consider the decrease in loss at each iteration, which can be expressed as:
\begin{align}
	& G_t(W_t) - G_{t+1}(W_{t+1}) =
	\nonumber \\
	& G_t(W_t) - G_t(W_{t+1}) + G_t(W_{t+1}) - G_{t+1}(W_{t+1})
	\nonumber \geq \\
	& G_t(W_t) - G_t(W_{t+1}) 
	\nonumber - \\
	& - (32\lambda_{\text{topo}} B ) \sqrt{\eta (G_t(W_t) - G_{t+1}(W_{t+1}))}.
\end{align}

According to the algorithm, prior to termination, the decrease in loss satisfies $G_t(W_t) - G_{t+1}(W_{t+1}) \geq \epsilon$. To guarantee a minimum decrease of $\frac{\epsilon}{2}$, we require a step-size $\eta$ that fulfills the following condition:
\begin{equation}
	\eta \leq \frac{G_t(W_t) - G_{t+1}(W_{t+1})}{4096\lambda_{\text{topo}}^2 B^2}.
\end{equation}

By leveraging the fact that $G_t(W_t) - G_t(W_{t+1}) \geq \epsilon$ and combining it with the constraint $\eta \leq \frac{1}{2C_2}$, we can conclude that it suffices to choose:
\begin{equation}
	\eta \leq \min \left\{ \frac{1}{2\ell^2 + 10\lambda_{\text{topo}}B}, \frac{\epsilon}{4096\lambda_{\text{topo}}^2 B^2}\right\}. \notag
\end{equation}

This ensures that $G_t(W_t) - G_{t+1}(W_{t+1}) \geq \frac{\epsilon}{2}$ holds whenever the stopping criterion is not met. Consequently, the algorithm terminates within $\frac{2C_0}{\epsilon}$ iterations.

This concludes the proof.

\paragraph{Remark 1.} Theorem 1 shows that the optimization process of the topology-aware total loss function can be made to converge efficiently by choosing a suitable step size. This choice is influenced by two terms, both of which depend on the cardinality of $\overline{Dgm^*}$. The use of truncated persistent barcodes in $\text{PersLoss}$ keeps $B=k_0+k_1+k_2$ small, making it possible to select a reasonable step size. However, if $B$ grows linearly with the dataset size, finding an appropriate step size becomes impractical.

\subsection{D. ADDITIONAL EXPERIMENTS}
In this section, we present additional experiments and analyses that complement the main text. These supplementary materials are intended to provide deeper insights and enhance the reader's understanding of our work.

\subsubsection{D1. Setup of Hardware and Software}
Our experiments were conducted on a machine running Ubuntu 18.04 LTS. The hardware configuration is as follows: two Intel Xeon Gold 6133 processors, each operating at 2.50 GHz, providing a total of 80 cores with 2 threads per core. The GPU setup includes eight NVIDIA GeForce RTX 3090 units. Each experimental scenario was run once on each of the eight GPUs, and the average result (performance metrics \& memory usage) was taken. The system is equipped with 196 GB of DDR4 RAM and has a cache memory of 129 GB, along with 20 GB of swap space. The architecture is x86\_64 with a Little Endian byte order. Our software environment includes CUDA 11.8, Python 3.10, and PyTorch 2.1.1. We utilize the GUDHI \cite{maria2014gudhi} and TopologyLayer \cite{gabrielsson2020topology} libraries for persistent homology computations.

\subsubsection{D2. Hyperparameter Sensitivity Analysis}
To further investigate the effectiveness of our two innovations, LPVI for enhancing initial geometric coverage and $\text{PersLoss}$ for adding topological constraints during training, we conduct a hyperparameter sensitivity analysis for both methods.

\textbf{(1) Hyperparameter of LPVI.}
LPVI has three hyperparameters: the number of neighbors in the neighborhood guiding 3D Voronoi interpolation ($\text{K\_max}$), the number of neighbors in the neighborhood guiding 2D Voronoi interpolation ($\text{K\_min}$), and a threshold value ($\text{Threshold}$) that measures the magnitude of local topological changes. If the persistence diagram distance before and after 3D Voronoi interpolation exceeds this threshold, it indicates that the local topological structure has been disrupted, and we switch to 2D Voronoi interpolation. Otherwise, we only perform 3D Voronoi interpolation. Since 3D Voronoi interpolation is more likely to alter the local topological structure, we set a larger number of neighbors in its neighborhood. In contrast, a smaller number of neighbors is sufficient to determine a hyperplane in 2D, so we set a smaller number of neighbors for 2D Voronoi interpolation. Therefore, we have $\text{K\_max} \geq \text{K\_min}$.

\begin{figure}[t]
	\centering
	\includegraphics[width=7.8cm]{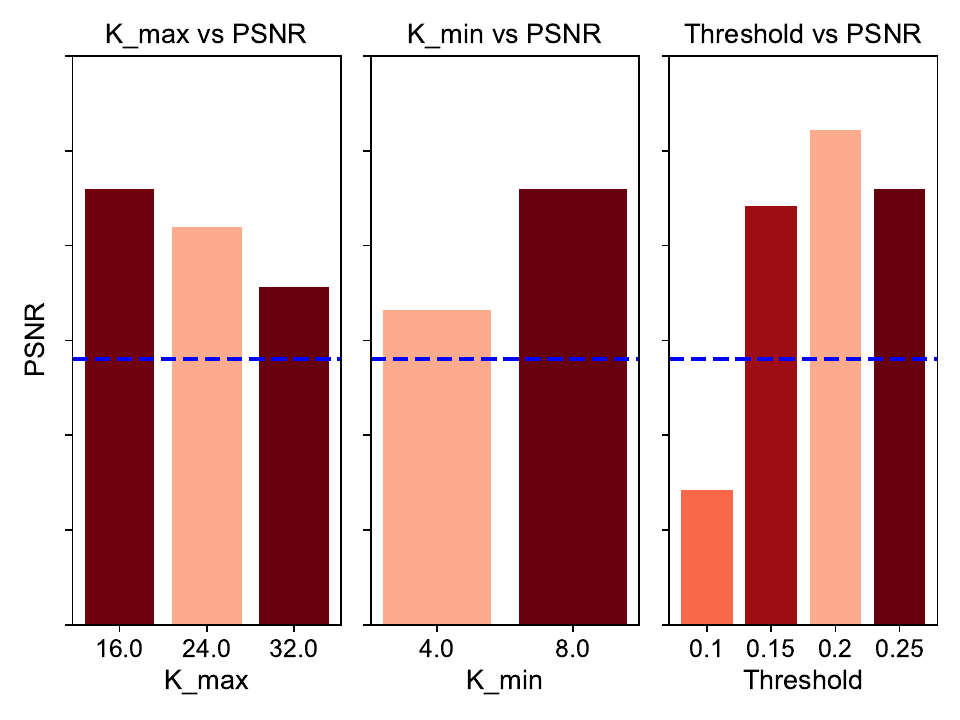}
	\caption{Hyperparameter sensitivity analysis of LPVI on Tanks \& Temples \cite{knapitsch2017tanks}. The PSNR metric corresponding to different hyperparameter values is presented in a bar chart, with the baseline PSNR marked by a blue dashed line. The color depth of the bars represents the memory overhead, with darker colors indicating higher memory usage.}
	\label{fig_abla_lpvi}
\end{figure}

From Figure \ref{fig_abla_lpvi}, we can observe that increasing $\text{K\_max}$ does not effectively improve performance. This is because a too-large neighborhood can lead to inter-pollution of local topological structures between different 3D Voronoi interpolations. Moreover, a smaller $\text{K\_max}$ does not necessarily result in reduced memory usage, as it may lead to more regions participating in 2D Voronoi interpolation. Similarly, $\text{Threshold}$ also requires a moderate value to achieve the best performance. The results suggest that a larger value of $\text{K\_min}$ tends to perform better among the tested values. This sensitivity analysis experiment yielded similar results on the Mip-NeRF360 \cite{barron2022mip} and Deep Blending \cite{hedman2018deep} datasets, which are not shown here for brevity.

\textbf{(2) Hyperparameter of $\text{PersLoss}$}
$\text{PersLoss}$ has four hyperparameters: the coefficient $\lambda_{\text{topo}}$ (abbreviated as $\text{lambda}$ in this section) that controls the influence of $\text{PersLoss}$ in the topology-aware total loss, and the parameters $k_0$, $k_1$, and $k_2$ of the Truncated Persistent Barcode, which represent the number of persistence points retained for 1D, 2D, and 3D homology, respectively.
The value of $\text{lambda}$ affects the magnitude of the impact of $\text{PersLoss}$ on the topology-aware total loss, while the values of $k_0$, $k_1$, and $k_2$ determine which dimensional topological features are effective. Generally, the feature maps used to compute LPIPS reflect the semantic features of the rendered image, and these abstract features are more suitable for matching with high-dimensional topological features.

\begin{figure}[t]
	\centering
	\includegraphics[width=7.8cm]{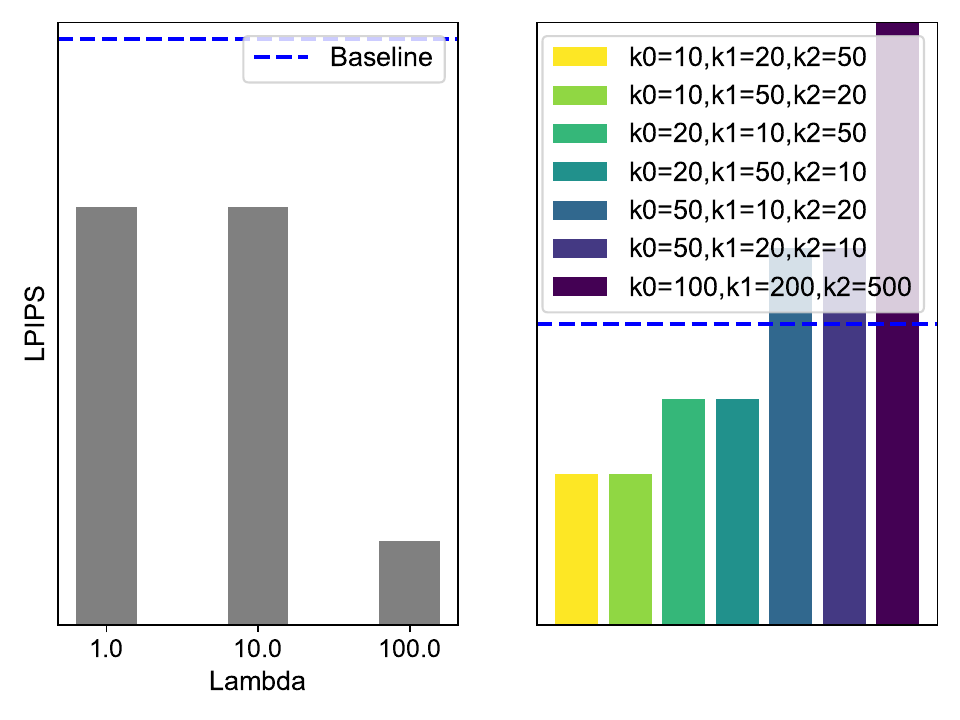}
	\caption{Hyperparameter sensitivity analysis of $\text{PersLoss}$ on Deep Blending \cite{hedman2018deep}. The LPIPS metric corresponding to different hyperparameter values is presented in a bar chart, with the baseline LPIPS marked by a blue dashed line.}
	\label{fig_abla_persloss}
\end{figure}

As illustrated in Figure \ref{fig_abla_persloss}, a larger value of $\text{lambda}$ is observed to lead to a decrease in the LPIPS metric. This can be attributed to the enhanced topological constraint imposed by $\text{PersLoss}$ in the topology-aware total loss function. A comparative analysis of the effects of $k_0$, $k_1$, and $k_2$ reveals that an increase in $k_0$ results in a corresponding increase in the LPIPS metric, suggesting that excessive emphasis on low-dimensional homology does not effectively guide the learning of semantic information in feature maps. In contrast, adjustments to $k_1$ and $k_2$ appear to have a negligible impact on the LPIPS metric, implying that these may be the desirable topological feature dimensions to learn. Finally, if $k_0$, $k_1$, and $k_2$ are all increased by an order of magnitude, $\text{PersLoss}$ becomes akin to a traditional topological loss function, retaining a large number of noise points. This noise-driven learning process leads to incorrect topological constraints during the optimization of the Gaussians, ultimately resulting in a significant increase in the LPIPS metric. This sensitivity analysis experiment yielded similar results on the Mip-NeRF360 \cite{barron2022mip} and Tanks\&Temples \cite{knapitsch2017tanks} datasets, which are not shown here for brevity.

\subsubsection{D3. Extended Experiments on Additional Datasets}
We further evaluated our model on more diverse and challenging datasets: all scenes from NeRF\_Synthetic \cite{mildenhall2021nerf}, six scenes from BungeeNeRF \cite{xiangli2022bungeenerf}, and four scenes from IMW2020 \cite{jin2021image}.
\begin{figure}[h]
	\centering
	\includegraphics[width=8.3cm]{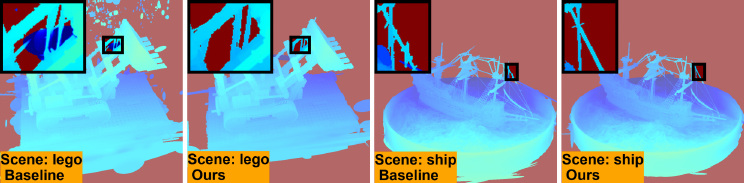}
	\caption{Visualization of depth maps from the Synthetic \cite{mildenhall2021nerf} dataset, showcasing two of the most complex scenes: lego and ship. Columns 1 and 3 display results from our baseline, i.e. Scaffold-GS, while columns 2 and 4 show results from the baseline with our $\text{PersLoss}$ regularization.}
	\label{fig_synthetic}
\end{figure}

\textbf{(1) Experiments on NeRF\_Synthetic}
We conducted experiments on the Synthetic dataset to validate our second innovation, namely, the incorporation of $\text{PersLoss}$ into the optimization process during the ADC stage. This additional constraint guides the training of Gaussians to learn structural information from the scene. As illustrated in Figure \ref{fig_synthetic}, the results demonstrate that this structural integrity is clearly reflected in the depth maps of the rendered 2D images.

The baseline method's depth map exhibits sluggish color transitions, failing to adapt to the sharp depth variations caused by structural elements (e.g., the mechanical arm of the Lego excavator and the mast of the ship). Furthermore, in the upper background of the Lego scene and the lower background of the Ship scene, the baseline method yields a larger number of isolated Gaussians that are not optimized properly, which can be attributed to the lack of topological constraints. Our approach, on the other hand, performs better, and this feature-level structural integrity is reflected in Table \ref{tab_synthetic}, where our method achieves a lower LPIPS score.

\begin{table}[h]
	\caption{LPIPS COMPARISON ON SYNTHETIC BENCHMARK}
	\label{tab_synthetic}
	\centering
	\resizebox{1\linewidth}{!} 
	{
		\begin{tabular}{c|cccccccc}
			& Chair                        & Drums                        & Ficus                        & Hotdog                       & Lego                         & Materials                    & Mic                          & Ship                         \\ \hline
			{\color[HTML]{656565} Baseline} & {\color[HTML]{656565} 0.013} & {\color[HTML]{656565} 0.047} & {\color[HTML]{656565} 0.014} & {\color[HTML]{656565} 0.022} & {\color[HTML]{656565} 0.018} & {\color[HTML]{656565} 0.041} & {\color[HTML]{656565} 0.008} & {\color[HTML]{656565} 0.113} \\
			Baseline+PersLoss               & 0.011                        & 0.044                        & 0.013                        & 0.021                        & 0.016                        & 0.040                        & 0.007                        & 0.110                       
		\end{tabular}
	}
\end{table}
The Synthetic dataset's scene structures are relatively simple, typically comprising only a single or a few basic foreground objects. As a result, the effect of $\text{PersLoss}$ on the LPIPS metric is limited. This is reflected in the feature maps, where both the baseline and our method's rendered images exhibit similar feature maps that closely match the ground truth. Furthermore, we did not assess the effectiveness of our first innovation, LPVI, which improves the geometric coverage of the initial point cloud output by SfM, on this benchmark. This is because the Synthetic dataset is a simulated dataset with uniformly distributed and dense initial point clouds within a cube, thereby eliminating the need for LPVI-based enhancements.

\begin{figure}[h]
	\centering
	\includegraphics[width=8.3cm]{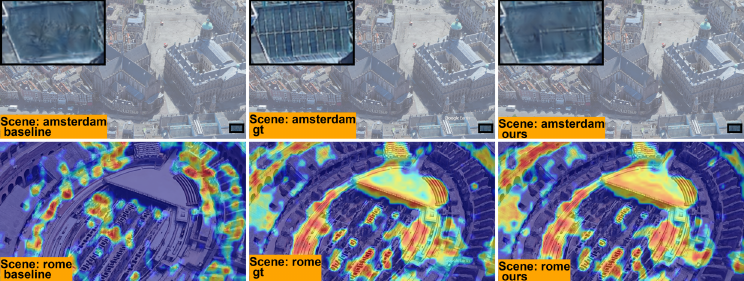}
	\caption{Visualization of results from the BungeeNeRF \cite{xiangli2022bungeenerf} dataset, specifically rendered images from the Amsterdam scene, and feature maps from the Rome scene. The columns show: (1) results from our baseline, i.e., Scaffold-GS; (2) ground-truth results; and (3) results from the baseline with our LPVI or $\text{PersLoss}$ regularization.}
	\label{fig_bungeenerf}
\end{figure}

\textbf{(2) Experiments on BungeeNeRF}
Following the evaluation protocol in Scaffold-GS, we also conducted experiments on the BungeeNeRF benchmark to test view-adaptive rendering. This dataset provides multi-scale outdoor observations and exhibits highly complex scene structures, making it an ideal testbed for evaluating the effectiveness of our approach in handling intricate and diverse environments.

\begin{table*}[t]
	\caption{COMPARISON OF DEPTH DISTORTION SOTA METHODS AND REAL-TIME PERFORMANCE}
	\label{tab_depth_and_realtime}
	\centering
	\resizebox{0.9\linewidth}{!} 
	{
		\begin{tabular}{ccccccc}
			\hline
			Method                 & \begin{tabular}[c]{@{}c@{}}Training Minutes\\ ↓\end{tabular} & \begin{tabular}[c]{@{}c@{}}Render FPS\\ ↑\end{tabular} & \begin{tabular}[c]{@{}c@{}}Indoor PSNR\\ ↑\end{tabular} & \begin{tabular}[c]{@{}c@{}}Indoor LPIPS\\ ↓\end{tabular} & \begin{tabular}[c]{@{}c@{}}Outdoor PSNR\\ ↑\end{tabular} & \begin{tabular}[c]{@{}c@{}}Outdoor LPIPS\\ ↓\end{tabular} \\ \hline
			2D-GS \cite{huang20242d}                   & \textbf{34}                                                  & 56                                                     & 30.39                                                   & 0.182                                                    & 24.33                                                    & 0.284                                                     \\
			GOF \cite{yu2024gaussian}                    & 205                                                          & 8                                                      & 30.79                                                   & 0.184                                                    & \textbf{24.82}                                           & \textbf{0.202}                                            \\
			Scaffold-GS \cite{lu2024scaffold} (Baseline) & 68                                                           & \textbf{85}                                            & 31.72                                                   & 0.167                                                    & 24.71                                                    & 0.269                                                     \\ \hline
			Baseline+LPVI (\textbf{ours})                  & 74                                                           & 75                                                     & \textbf{31.79}                                          & 0.166                                                    & 24.76                                                    & 0.268                                                     \\
			Baseline+PersLoss (\textbf{ours})              & 90                                                           & 84                                                     & 31.75                                                   & \textbf{0.164}                                           & 24.73                                                    & 0.267                                                     \\ \hline
		\end{tabular}
	}
	\begin{flushleft}
		\footnotesize
		* \textbf{Bold} indicates the best results.
	\end{flushleft}
\end{table*}

\begin{table}[h]
	\caption{PSNR AND LPIPS COMPARISON ON BUNGEENERF BENCHMARK}
	\label{tab_bungeenerf}
	\centering
	\resizebox{1\linewidth}{!} 
	{
		\begin{tabular}{c|cccccc}
			& Amsterdam                    & Bilbao                       & Hollywood                    & Pompidou                     & Quebec                       & Rome                         \\ \hline
			{\color[HTML]{656565} Baseline (PSNR)}  & {\color[HTML]{656565} 28.06} & {\color[HTML]{656565} 29.28} & {\color[HTML]{656565} 26.38} & {\color[HTML]{656565} 27.08} & {\color[HTML]{656565} 28.84} & {\color[HTML]{656565} 28.03} \\
			Baseline+LPVI (PSNR)                    & 28.30                        & 29.68                        & 26.82                        & 27.66                        & 29.27                        & 28.65                        \\
			{\color[HTML]{656565} Baseline (LPIPS)} & {\color[HTML]{656565} 0.098} & {\color[HTML]{656565} 0.099} & {\color[HTML]{656565} 0.159} & {\color[HTML]{656565} 0.099} & {\color[HTML]{656565} 0.097} & {\color[HTML]{656565} 0.099} \\
			Baseline+PersLoss (LPIPS)               & 0.089                        & 0.093                        & 0.145                        & 0.095                        & 0.093                        & 0.084                       
		\end{tabular}
	}
\end{table}

From the comparison of the first row of images in Figure \ref{fig_bungeenerf}, it can be observed that our proposed LPVI method is effective in interpolating low-curvature regions even in complex scenes with initial point clouds, ultimately enabling the rendering of texture on planar object surfaces in the rendered image and improving pixel-level structural integrity. The comparison of the second row of images in Figure \ref{fig_bungeenerf} reveals that our designed $\text{PersLoss}$ guides the rendered image to exhibit semantic features similar to the ground truth, as reflected in the feature maps, which are closer to the ground truth feature maps compared to the baseline, indicating an enhancement of feature-level structural integrity.
Furthermore, Table \ref{tab_bungeenerf} provides a quantitative demonstration of the improvement in pixel-level structural integrity, as evidenced by the increase in PSNR, and the improvement in feature-level structural integrity, as indicated by the decrease in LPIPS.

\textbf{(3) Experiments on IMW2020}
To demonstrate the versatility of our approach and its potential to benefit 3D-GS in other tasks, we conducted experiments on the image matching benchmark. This real-world dataset poses a significant challenge, as the images were captured in the presence of numerous dynamic objects, such as people, which can occlude the target object from certain viewpoints \cite{yuan2025sed}. Consequently, our method's ability to understand the scene's structure is crucial for effective reconstruction. 

\begin{table}[h]
	\caption{PSNR AND LPIPS COMPARISON ON IMW2020 BENCHMARK}
	\label{tab_imw2020}
	\centering
	\resizebox{1\linewidth}{!} 
	{
		\begin{tabular}{c|cccc}
			& brandenburg\_gate            & buckingham\_palace           & colosseum\_exterior          & florence\_cathedral\_side    \\ \hline
			{\color[HTML]{656565} Baseline (PSNR)}  & {\color[HTML]{656565} 15.09} & {\color[HTML]{656565} 14.78} & {\color[HTML]{656565} 13.49} & {\color[HTML]{656565} 12.80} \\
			Baseline+LPVI (PSNR)                    & 15.33                        & 15.00                        & 13.73                        & 13.04                        \\
			{\color[HTML]{656565} Baseline (LPIPS)} & {\color[HTML]{656565} 0.345} & {\color[HTML]{656565} 0.347} & {\color[HTML]{656565} 0.350} & {\color[HTML]{656565} 0.329} \\
			Baseline+PersLoss (LPIPS)               & 0.343                        & 0.346                        & 0.345                        & 0.322                       
		\end{tabular}
	}
\end{table}

We present the PSNR and LPIPS metrics in Table \ref{tab_imw2020}. Notably, the LPIPS metric shows limited improvement, which can be attributed to the fact that the appearance and disappearance of moving objects can cause the Gaussians in 3D space to learn biased structural information, resulting in artifacts in the rendered images. These artifacts fundamentally reflect that the Gaussians have not converged to a good structural state, and it is challenging to recover from this solely relying on the topological constraints imposed by $\text{PersLoss}$. In contrast, the PSNR metric shows some improvement at the pixel level, indicating that our method has a positive effect on the rendering quality. Future work may focus on developing dynamic topological constraints that can adapt to changing scenes, enabling more effective reconstruction of dynamic environments.

\subsubsection{D4. Comparisons with Depth Distortion SOTA Methods}
We compare our method with SOTA approaches for addressing depth distortion, including 2D-GS \cite{huang20242d} and GOF \cite{yu2024gaussian}. While all methods aim to enhance 3D geometry reconstruction, their approaches differ fundamentally. 2D-GS and GOF directly constrain geometry by aligning Gaussians along rays using depth information. In contrast, our method leverages a topological perspective, using PH to compute differences between rendered 2D images and GT images. These differences indirectly constrain Gaussians by propagating gradients back into 3D space.
As shown in Table \ref{tab_depth_and_realtime}, our method outperforms 2D-GS and GOF across key metrics, achieving higher PSNR and lower LPIPS in both indoor and outdoor settings. This improvement is attributed to the effective use of accurate GT images for 2D rendered results, enabling more robust optimization.

\subsubsection{D5. Real-Time Analysis}
Table \ref{tab_depth_and_realtime} presents the real-time performance of our method, evaluated using two primary metrics: \textit{training time}, defined as the duration required for 3D scene reconstruction, and \textit{rendering speed}, measured in frames per second (FPS) for novel-view synthesis.

The LPVI module introduces an additional overhead of approximately 6 minutes per scene (without parallelism) due to PH computations. PersLoss further increases training time during the ADC stage (epochs 1500–15000, triggered every 200 iterations), with each computation taking about 20 seconds. For rendering speed, the LPVI module slightly reduces FPS because of the optimization of additional Gaussians. However, since PersLoss is applied only during training, it does not affect inference speed. Despite these computational costs, both the LPVI module and PersLoss significantly enhance rendering quality, as demonstrated by the evaluation metrics in Table \ref{tab_depth_and_realtime}.

To enhance runtime performance, LPVI and PH computations can be precomputed and cached for static scenes. For dynamic scenes, efficiency could be improved by parallelizing the LPVI module or adopting faster PH algorithms. These strategies highlight the scalability and adaptability of our method, paving the way for future advancements in real-time performance and rendering quality.

\end{document}